%% file: main.tex
\documentclass[11pt,a4paper]{article}
\usepackage{times,latexsym}
\usepackage{url}
\usepackage[T1]{fontenc}

\usepackage[acceptedWithA]{tacl2021v1}

\usepackage{xspace,mfirstuc,tabulary}

\newif\iftaclinstructions
\taclinstructionsfalse 
\iftaclinstructions
\renewcommand{\confidential}{}
\renewcommand{\anonsubtext}{(No author info supplied here, for consistency with
TACL-submission anonymization requirements)}
\newcommand{\instr}
\fi

\iftaclpubformat 

\else

\fi

\usepackage{graphicx}
\usepackage{booktabs}
\usepackage{multirow}
\usepackage{censor}
\usepackage{amsmath}
\usepackage{amssymb}
\usepackage{dblfloatfix}
\usepackage{enumitem}
\usepackage{dialogue}
\usepackage{tcolorbox}
\usepackage{hyperref}
\usepackage{textcomp}  
\usepackage{scalerel}  
\newtheorem{definition}{Definition}
\def\pauseemoji{{\includegraphics[scale=0.3]{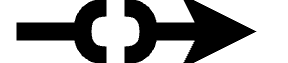}}}
\def\specifyemoji{{\includegraphics[scale=0.3]{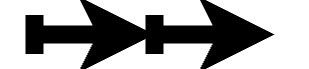}}}
\def\revealemoji{\scalerel*{\includegraphics{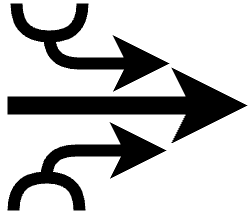}}
{\textrm{\textbigcircle}}}
\def\huskyemoji{\scalerel*{\includegraphics{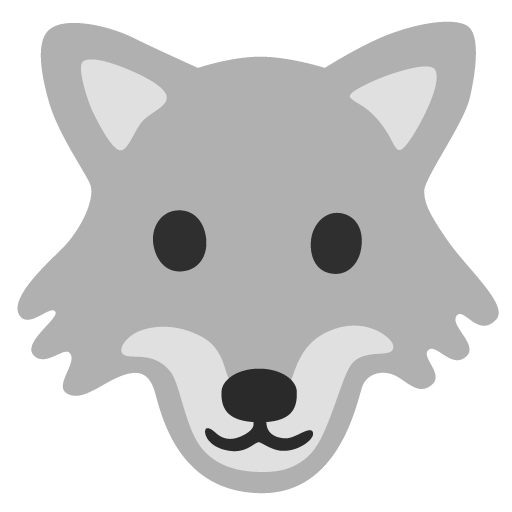}}
{\textrm{\textbigcircle}}}
\def\uscemoji{\scalerel*{\includegraphics{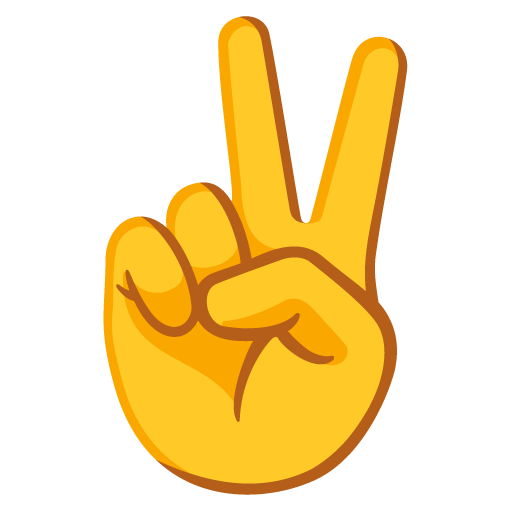}}
{\textrm{\textbigcircle}}}
\def\kingfisheremoji{\scalerel*{\includegraphics{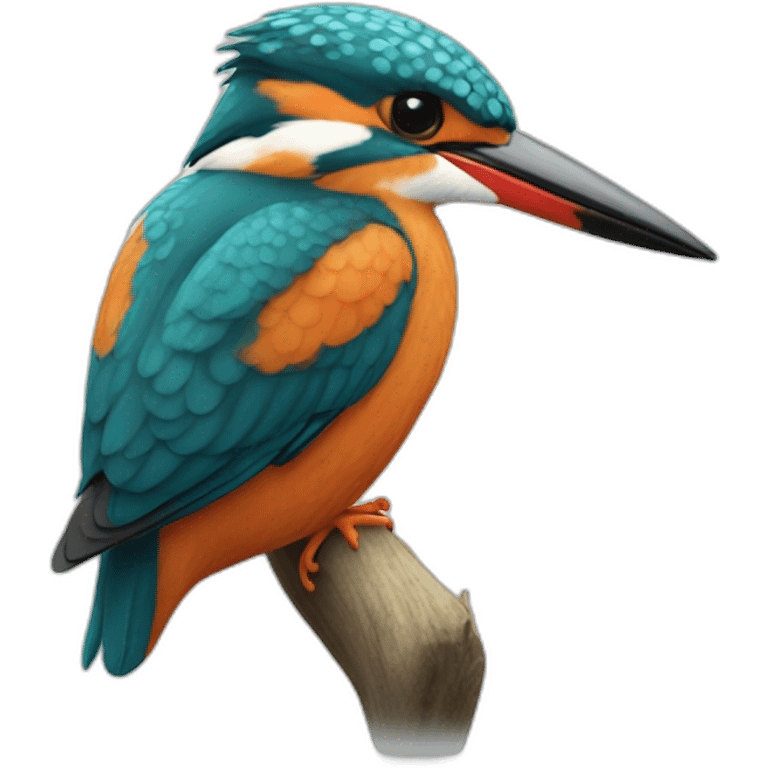}}
{\textrm{\textbigcircle}}}

\title{Better \textit{Slow} than Sorry:\\Introducing Positive Friction for Reliable Dialogue Systems}

\author{
    Mert İnan\textsuperscript{\rm \huskyemoji},
    Anthony Sicilia\textsuperscript{\rm \huskyemoji}, 
    Suvodip Dey\textsuperscript{\rm \kingfisheremoji}, 
    Vardhan Dongre\textsuperscript{\rm \kingfisheremoji}, 
    Tejas Srinivasan\textsuperscript{\rm \uscemoji}\\
    \textbf{Jesse Thomason}\textsuperscript{\rm \uscemoji},
    \textbf{Gökhan Tür}\textsuperscript{\rm \kingfisheremoji},
    \textbf{Dilek Hakkani-Tür}\textsuperscript{\rm \kingfisheremoji},
    \textbf{Malihe Alikhani}\textsuperscript{\rm \huskyemoji} \\
    \textsuperscript{\rm \uscemoji} University of Southern California \ \ 
    \textsuperscript{\rm \kingfisheremoji} University of Illinois Urbana-Champaign \\
    \textsuperscript{\rm \huskyemoji} Northeastearn University \\
    \texttt{\{inan.m, alikhani.m\}@northeastern.edu}
}

\date{}

\begin{document}
\maketitle
\begin{abstract}
While theories of discourse and cognitive science have long recognized the value of unhurried pacing, recent dialogue research tends to minimize friction in conversational systems. Yet, frictionless dialogue risks fostering uncritical reliance on AI outputs, which can obscure implicit assumptions and lead to unintended consequences. To meet this challenge, we propose integrating \textit{positive friction} into conversational AI, which promotes user reflection on goals, critical thinking on system response, and subsequent re-conditioning of AI systems. We hypothesize systems can improve goal alignment, modeling of user mental states, and task success by deliberately slowing down conversations in strategic moments to ask questions, reveal assumptions, or pause. We present an ontology of positive friction and collect expert human annotations on multi-domain and embodied goal-oriented corpora. Experiments on these corpora, along with simulated interactions using state-of-the-art systems, suggest incorporating friction not only fosters accountable decision-making, but also enhances machine understanding of user beliefs and goals, and increases task success rates.\footnote{Code, data, and guidelines will be made public.}%
\end{abstract}

\input{sections_new/01_intro}
\input{sections_new/02_related_work}
\input{sections_new/03_positive_friction}

\input{sections_new/04_analyzing_friction}
\input{sections_new/05_applying_friction}
\input{sections_new/06_discussion}
\input{sections_new/07_trailing_sections}

\bibliography{custom, tacl2021}
\bibliographystyle{acl_natbib}

\onecolumn

\appendix
\input{sections_new/xx_appendix}

\end{document}

%% file: sections_new/01_intro.tex
\begin{figure}[!h]
    \centering
    \includegraphics[width=\columnwidth]{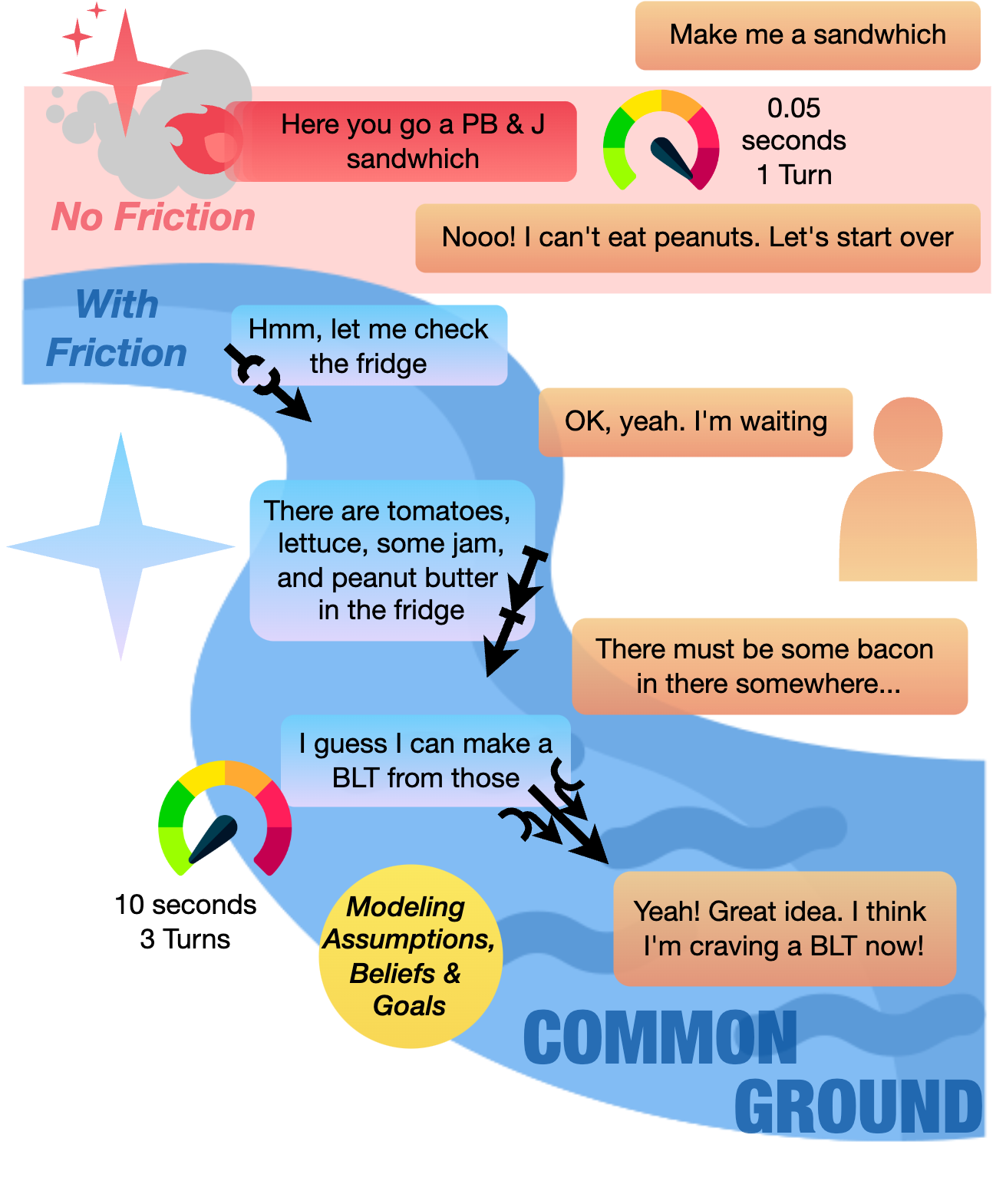}
    \vspace{-30pt}
    \caption{We characterize \textit{positive friction} in dialogues to better model user goals, beliefs, and assumptions. This paper introduces an ontology of positive friction ``movements'' such as reflective pausing (\pauseemoji), overspecification (\specifyemoji), or assumption reveal (\revealemoji). We show that frictive conversations increase user satisfaction and task success, despite creating longer dialogues.}
    \label{fig:main_fig}
\vspace{-0.1in}
\end{figure}

\section{Introduction}
Theories of common ground reveal how the rhythm and timing of dialogue shape the dynamics of interaction, fostering clarity and mutual understanding \cite{Stalnaker1978Dec, tannen1989talking, Wilkes-Gibbs1992Apr, zellner1994pauses}. They claim productive interactions involve a dynamic interplay of questioning, challenging assumptions, disclosing beliefs, and offering elaborations---actions that momentarily slow the dialogue but ultimately enhance reasoning and collaboration. Meanwhile, current LLM-based conversational systems are trained on user preferences, conflating superficial sentiment with the nuanced, underlying sub-goals of communication. This strategy biases systems toward specific length preferences \citep{Geishauser2024Apr, zhang-etal-2024-strength} without considering the long-term value of individual utterances.

To resolve this issue, we argue that conversational systems should incorporate deliberate moments of \textit{positive friction}---movements that decelerate the dialogue to reveal the underlying goals and assumptions of both interlocutors. 
To motivate this position, we present an ontology to characterize communicative acts that incorporate positive friction. We study this ontology and its utility in goal-oriented collaborations between humans and AI systems.
This approach models nuanced conversational actions and opens new avenues for evaluating dialogue systems. 
Our work advocates for a shift in dialogue system design, prioritizing long-term collaboration over short-term efficiency to build more reliable systems in terms of both user interfacing and system response.

Our argument is rooted in the perspective that utterances in a dialogue hold different \textit{valence} -- the impact of an utterance on dialogue speed. Indirectly, modifying dialogue speed adjusts conversational flow to make time for improved common ground and modeling of user mental states. For instance, in Figure~\ref{fig:main_fig}, pausing to say ``Hmm, let me check the fridge'' redirects the course of the interaction to a valuable outcome previously unknown to the user. Current dialogue systems may not accomplish this because both dialogue management policies \cite{li-etal-2016-deep, li-etal-2017-end} and evaluation frameworks \cite{liu-etal-2016-evaluate, li2021evaluatedialoguemodelsreview, braggaar2024evaluatingtaskorienteddialoguesystems} favor \textit{frictionless} and \textit{efficient} conversations (\textit{e.g.}, penalizing each additional conversational turn). Addressing this disparity in valence, our positive friction ontology has significant implications for shaping reward policies in dialogue management systems and evaluation metrics.

In line with the above argument, we ask,
\begin{enumerate}[nolistsep,leftmargin=1.5em]
    \item \textit{What counts as positive friction}?
    \item \textit{Does friction improve modeling of user goals}?
    \item \textit{Does this equate to improved task success}?
\end{enumerate}
To answer these questions, we introduce the concept of positive friction in relation to goal-oriented conversations (\S \ref{sec:pos_fric}). 
We develop a novel multimodal taxonomy that integrates cognitive and linguistic theories of discourse to classify various types of frictive movements (\S\ref{subsec:taxonomy}), and collect human annotations for two tasks: detecting and generating friction movements (\S \ref{sec:data_collection}). 
We further highlight the relationship between friction and traditional dialogue acts in three conversation datasets (\S \ref{sec:comparative_analysis}): MultiWOZ, TEACh, and PersuasionForGood. 
We demonstrate the benefits of positive friction in real-time task-oriented dialogues, finding improvement in both modeling of user mental states~ (\S\ref{sec:analyze_friction}) and user goals while requiring fewer overall actions (\S \ref{sec:applying_friction}). 
We conclude by discussing new methodologies for dialogue evaluation through the lens of positive friction (\S \ref{sec:discussion}). 

%% file: sections_new/02_related_work.tex
\section{Related Work}

\begin{figure*}
    \centering
    \includegraphics[width=0.92\linewidth]{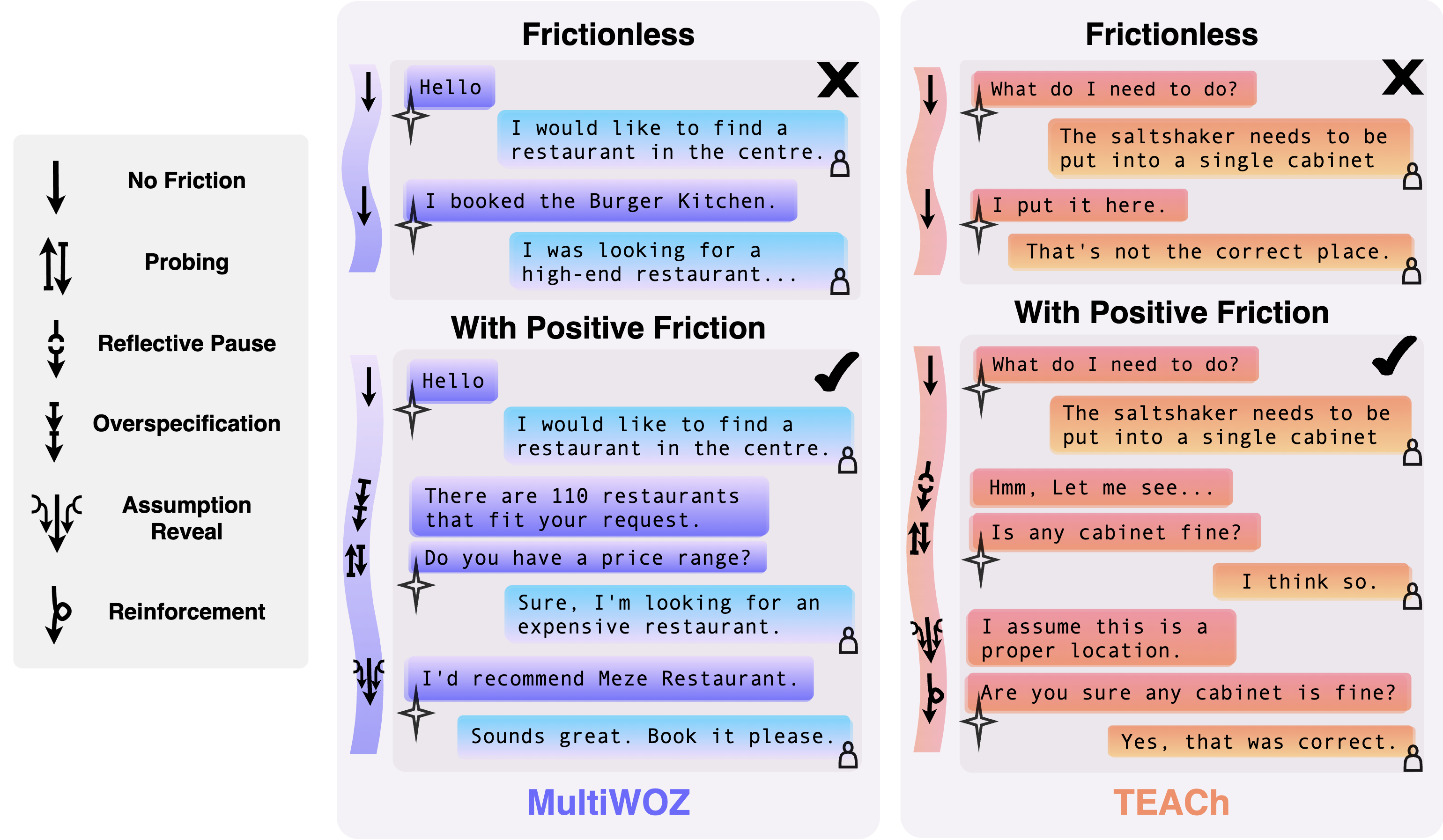}
    \caption{A comparative example of conversations based on TEACh and MultiWOZ datasets. Frictionless conversations take fewer turns, but may not result in successful completion of the task given by the user. Conversations with multiple positive friction movements lead to longer but ultimately more successful conversations.}
    \label{fig:friction_dialogue_example}
\vspace{-0.1in}
\end{figure*}

Slowing down user experiences through friction is typically viewed as undesirable, due to the risk of increased user frustration. 
However, frictionless experiences on social media \cite{Anderson2021Jan}, smartphone apps \cite{lee2010indagator}, and online platforms \cite{Lembke_2023} that exploit the brain's short-term reward mechanisms~\cite{fishbein2011predicting} could be detrimental to users' long-term goals~\cite{ericson2022reimagining}, especially increasing sycophancy when uncertain \cite{sicilia2024accounting}. 
To promote reflective interactions that are more beneficial in the long term, we propose incorporating \emph{positive friction} into dialogues.

Cognitive science has studied how the human brain employs different neural systems for short- and long-term goals~\cite{Evans2003Oct, McClure2007May, Diekhof2010Jan}. 
Friction occurs at the interface of these two systems, by moving the user away from instinctive System-1 thinking into reflective System-2 thinking. 
Subsequently, design philosophies such as slow technology~\cite{Hallnas2001Aug} and undesign thinking~\cite{pierce2014undesigning} promote reflection by \emph{intentionally} designing friction into user experience. 
Various works have explored the potential of slowing down interactions to reduce errors~\cite{10.1145/1753846.1754054, brumby2013recovering, soboczenski2013increasing}, disrupt ``mindless'' interactions~\cite{Cox2016May, ruiz2024design}, challenge users~\cite{cairns2014immersion} and promote behavior change~\cite{10.1145/2702123.2702537}. 
These examples show the importance of technological designs that promote reflective user thinking.

Similar benefits have been observed when human-AI interactions are deliberately decelerated. 
Abstention~\cite{de2000reject} and deferral~\cite{mozannar2023should, lemmer2023human} improve AI reliability under uncertainty, while nudges~\cite{caraban201923}, epistemic markers~\cite{zhou-etal-2024-relying, kim2024m} and cognitive forcing functions~\cite{buccinca2021trust, park2019slow, ma2024you} can mitigate over-reliance. 
In embodied and multimodal settings, employing friction to build more common ground facilitates better human-machine interaction~\cite{Marge2013, DBLP:journals/aim/ChaiFLS16, Hough2017Mar, Carlmeyer2018Mar, sicilia2023isabel, atwell2024combining}. 

However, these frictive behaviors are not naturally built into modern LLM-based dialogue systems. 
\citet{zhou-etal-2024-relying} find that LLMs rarely express uncertainty, even when they are incorrect. 
They trace this behavior to RLHF preference datasets, where LLM outputs with uncertainty expressions were usually rejected by human raters. 
Further, LLMs' tendency for sycophancy~\cite{sharma2023towards, malmqvist2024sycophancy} 
and hallucination instead of abstaining~\cite{huang2023survey} are further indications of their aversion to introducing friction. 
Fostering mindful interactions with LLM-based systems requires intentional inclusion of friction~\cite{collins2024modulating}. 
To facilitate this goal, we draw on theories of communication and discourse~\cite{Stalnaker1978Dec, tannen1989talking, Wilkes-Gibbs1992Apr} and develop a novel taxonomy that codifies friction in goal-oriented dialogue. 
Our taxonomy provides a scaffolding for designing training and inference methods that enable LLMs to engage users in a collaborative, reflective reasoning process, leading to more reliable and satisfying human-AI interactions.

%% file: sections_new/03_positive_friction.tex
\input{sections_new/table_frictionexamples}

\section{Positive Friction in Dialogue}
\label{sec:pos_fric}
We introduce the concept of positive friction for goal-oriented dialogue, and our taxonomy that captures different types of friction movements (\S\ref{subsec:taxonomy}). 
We further use our taxonomy's categories to annotate two dialogue datasets (\S\ref{sec:data_collection}), and discuss their relationship to dialogue acts (\S\ref{sec:comparative_analysis}).

\begin{definition}[Positive Friction]
Positive frictions are intentional movements that slow down the course of an interaction in order to yield positive long-term impact. These movements may not be strictly necessary for task completion, and may be perceived as intrusive or unwelcome by the user, but can encourage System-2 thinking \citep{Evans2003Oct} such as reflection on uncertain assumptions by both users and AI systems.\footnote{This definition is constructed using behavioral science literature~\cite{caraban201923, chen2024exploringbehavioralmodelpositive}.}
\end{definition}

Figure~\ref{fig:friction_dialogue_example} illustrates the differences between frictionless conversations and those with multiple frictions. In frictionless conversations, assumptions are not revealed, no questions are asked, and no reasoning is shared. 
The interaction is shorter but may result in undesirable outcomes (\textit{e.g.} a restaurant booking that does not fit the user's preferences). 
On the other hand, positive friction frequently stalls the conversation by asking questions and providing additional information and explanations, thereby encouraging incremental steering of the conversation.

We build on concepts from cognitive science and discourse literature to codify unique classes for different positive friction \textit{movements}, which can be used in addition to dialogue acts to better capture mind perception capabilities \cite{waytz2010causes} inherent to human conversations. 

\subsection{Taxonomy of Friction Movements}
\label{subsec:taxonomy}

We introduce a new taxonomy of positive friction movements that can change the course of an interaction with minimal short-term intrusions, resulting in a long-term positive outcome. We define several high-level categories and subcategories for different friction movements.

While high-level categories are based on prior dialogue theories introduced by linguistics and cognitive science literature, the sub-categories are based on a pragmatic classification approach, where the setting of the conversation is used to distinguish classes. 
Please refer to Table~\ref{tab:friction_examples} for examples of these classes.

\paragraph{Assumption Reveal:} The speaker reveals their subjective assumptions or beliefs about the environment, actions, or other interlocutors. Revealing these assumptions uncovers information previously hidden from one interlocutor (or implicitly assumed) and opens up new avenues for conversation. This category is based on belief coordination~\citep{Wilkes-Gibbs1992Apr}.
    \begin{itemize}[nolistsep, leftmargin=1em]
        \item \textbf{Contextual Assumption Reveal:} The speaker reveals assumptions about the environment.
        \item \textbf{Conversational Assumption Reveal:} The speaker reveals assumptions about previously mentioned utterances in the conversation. 
        \item \textbf{Metacognitive Assumption Reveal:} The speaker reveals their assumptions about their own or the other interlocutor's reasoning, plans or goals in the conversation. 
    \end{itemize}

\paragraph{Reflective Pause:}  The speaker pauses while producing an utterance or breaks their sentence to depict uncertainty, a sudden change in the environment, or a new action being taken. 
This movement is analogous to the pause types studied by \citet{zellner1994pauses, fors2015production, reed2017analysing}.
    \begin{itemize}[nolistsep, leftmargin=1em]
        \item \textbf{Conversational Pause:} Verbal or non-verbal cues that depict internal reflection.
        \item \textbf{Embodied Pause:} While interacting with the environment, the speaker intentionally pauses using their physical body. 
        \item \textbf{Recalibrating Pause:} When a change in plan occurs, the speaker intentionally pauses and changes the course of action. 
    \end{itemize}
\paragraph{Reinforcement:} The speaker restates their own previous utterance for emphasis, rewinding the flow of the conversation. 
This movement is similar to ``repetition in discourse''~\citep{tannen1989talking}. 

\paragraph{Overspecification:} 
The speaker relays additional, overly-specific information that was not requested, but may nevertheless be useful to the other interlocutor. 
This category is based on bounded-rational overspecification~\citep{Tourtouri2021Dec} which posits that humans are only moderately Gricean during conversations \cite{Mangold1988Sep, Engelhardt2006May}. 
    \begin{itemize}[nolistsep, leftmargin=1em]
        \item \textbf{Elaborative Overspecification:} The speaker gives more details, specificity, or additional explanation about their actions or the environment. This adds to the conversation what was already known by both interlocutors. 
        \item \textbf{Confirmative Overspecification:} The speaker confirms and elaborates the actions, choices, or beliefs. Examples include a repetition of previous utterances, elaborate responses to yes/no questions, or longer than necessary responses. 
    \end{itemize}

\paragraph{Probing:} The speaker poses a question regarding an external aspect of the conversation, such as the environment, the actions, or the interlocutors, redirecting the flow of the conversation to the other interlocutor. This movement is built on communal inquiry basis of discourse~\citep{Stalnaker1978Dec, Roberts2012Dec}.
\begin{itemize}[nolistsep, leftmargin=1em]
    \item \textbf{Contextual Probing:} The speaker asks a question regarding the environment, actions, or interlocutors in an effort to better understand the context and resolve ambiguities. 
    \item \textbf{Conversational Probing:} The speaker asks a question to clarify something previously mentioned in the conversation. 
    \item \textbf{Plan-level Probing:} The speaker asks a question regarding the goal, reasoning, or future steps in order to plan out their actions better. 
\end{itemize}

\vspace{0.5em}
Depending on the context and conversation to which these movements are applied, new subcategories can be introduced for each higher-level category. In the following data collection effort, we use these subcategories to validate this proposed hierarchy. In subsequent sections of the paper, we use only higher-level categories to investigate the applicability of positive friction as a concept.

\subsection{Human Annotation of Friction}
\label{sec:data_collection}
To test the empirical validity of our ontology, we collect human annotations on two collaborative, task-oriented dialogue datasets.

\paragraph{Data}
We use two datasets, MultiWOZ and TEACh, for the annotations. These datasets are chosen specifically due to their wide range of multi-domain/embodied tasks, allowing us to observe the role of positive friction. We extract all dialogues from both datasets. For the annotator's convenience, in TEACh dialogues, we only show textual utterances, not the interaction's video feed.

\paragraph{Annotation Protocol}
We hired 10 engineering undergraduate students and administered a short lecture on the linguistic and cognitive science background of positive friction in conversations---they are henceforth referred to as expert annotators. Next, expert annotators were presented with an annotation interface and asked to complete two tasks: detection and production. Participants were given an information sheet that outlined task details and the benefits and risks of participating in this study. Participation was voluntary and annotators were compensated \$15/hour.\footnote{This annotation protocol is approved by our institution's IRB committee. Out of the 10 participants, data from 1 participant was dropped due to validation errors. The annotation interface can be found in Appendix~\S\ref{sec:appendix_annotation_interface}.} In total, 430 minutes were spent on annotating 714 questions.

\begin{figure*}
    \includegraphics[width=\linewidth]{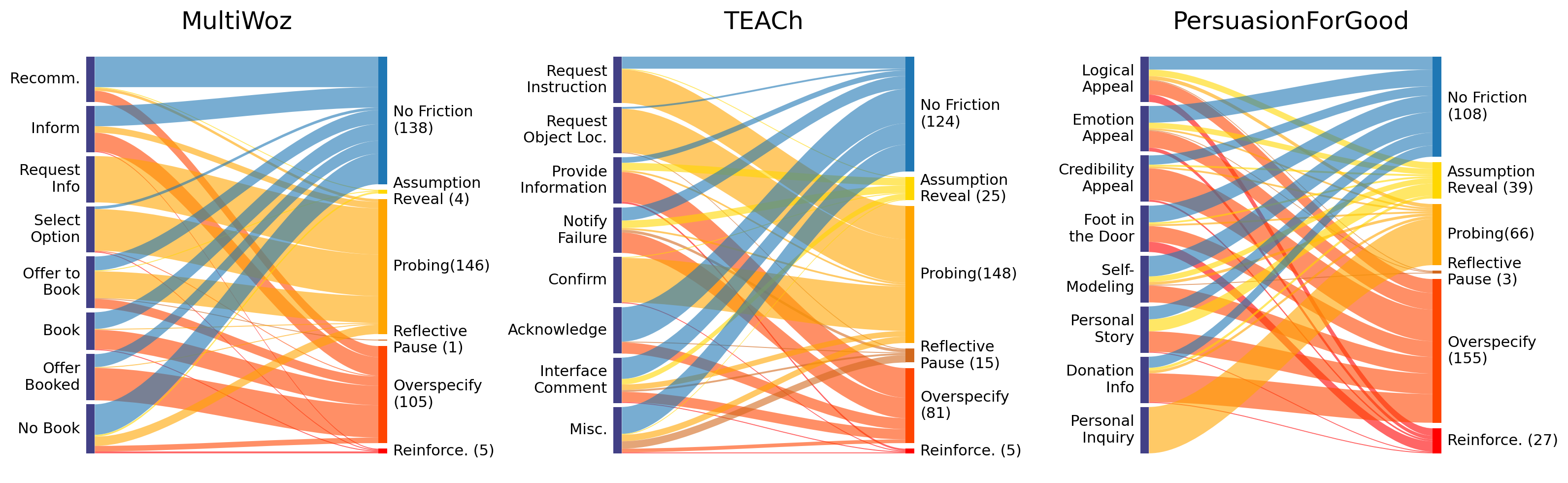}
    \caption{Distribution of 50 utterances sampled from annotated dialogue acts (left) belonging to three dialogue datasets into friction categories (right), as annotated by GPT-4o. 
    Most dialogue acts can occur both with and without friction. 
    For example, in TEACh, failure notifications may lack friction, reveal assumptions by suggesting alternatives, or overspecify failure details.
    }
    \label{fig:friction-dialogacts-distribution}
\vspace{-0.1in}
\end{figure*}

\paragraph{Task 1: Positive Friction Detection} 
Each annotator is provided with two randomly selected dialogues,  one from MultiWOZ and one from TEACh. 
For each turn in both dialogues, the annotator is asked to select the friction movement corresponding to that turn (or ``No friction'', if none applies). 
Additionally, for inter-annotator agreement calculations, all annotators annotate one common dialogue from each of MultiWOZ and TEACh. 
In total, for this task, 494 utterances were annotated with 284 distinct questions and 21 shared questions among annotators. On average, each annotator annotated $17 \pm 12$ turns of dialogue for TEACh and $12 \pm 5$ turns for MultiWOZ. 

We calculate inter-annotator agreement using Cohen's Kappa (heatmap in Appendix~\S\ref{sec:appendix_interannotator}). 
The highest agreement is around 0.42 for category-level and 0.38 for subcategory-level, which points to a fair to moderate level agreement. Instances of disagreement stem from the subjectivity of when to use friction. These kappa scores are comparable to other discourse-level annotation paradigms such as in the STAC ($\sim$0.5) \cite{asher-etal-2016-discourse} and Potsdam Commentary ($\sim$0.28) Corpora \cite{stede-neumann-2014-potsdam}.

\paragraph{Task 2: Positive Friction Production}
Annotators are provided with a partial dialogue until a randomly-selected turn (50\% user turns, and 50\% assistant turns). They are then asked to generate a reply to the last utterance containing a friction movement of their choice. 
They are allowed to select multiple friction movements and write an utterance for each, or write a reply on the "other" category if nothing is appropriate. 
Each annotator generates frictive responses for 20 randomly-selected partial dialogues each from MultiWOZ and TEACh. 
In total, 220 dialogues are given to all annotators, with 200 distinct dialogues, and 2 dialogues common among all annotators.

We find that people prefer to use questions to create friction most of the time (a result visible in the histograms found in the Appendix~\S\ref{sec:appendix_histogram}). 
Questions can be essential, but other forms of friction are also used. In fact, in real-world conversations, the histogram follows a Zipfian distribution \cite{Piantadosi2014Oct}, whereby there is a tailing end in the histogram that accounts for a significant amount of less common but important uses of other friction movements.

\subsection{Automatic Detection of Friction}
\label{subsec:automatic-detection}
In addition to the human annotations for positive friction detection, we investigate a simple automated approach for friction detection that uses LLMs for proxy annotation. 
We prompt \href{https://openai.com/gpt-4o-contributions/}{Open AI's GPT-4o}\footnote{gpt-4o-2024-08-06} with the same annotation manual provided to human annotators (as system prompt) and further prompt the model to determine a category-level friction movement. When compared to each human annotator, GPT-4o has an average Cohen's Kappa of about \textbf{0.34} across all turns and about \textbf{0.20} on the smaller subset of turns that all annotators shared. GPT-4o's agreement with the majority vote of all annotators is moderate at \textbf{0.50}.
Appendix~\S\ref{sec:appendix_in_out_gpt} contains a detailed analysis.

\subsection{Friction Categories Extend Dialogue Acts}
\label{sec:comparative_analysis}

We further study the connection between friction and other bottom-up ontological efforts, namely dialogue acts. 
Dialogue acts are categorizations of utterances that represent a specific intent. 
As such, it is natural to consider the relationship between friction and traditional dialogue act categories. 
Can the same dialogue act occur in both frictive and non-frictive forms?
Are some dialogue acts inherently frictive? 
Which friction categories are most commonly represented in existing dialogue datasets?

We focus on three conversational datasets that have dialogue act annotations: MultiWOZ~\cite{multiwoz}, TEACh~\cite{padmakumar2021teachtaskdrivenembodiedagents}, and PersuasionForGood~\cite{wang-etal-2019-persuasion}.\footnote{For PersuasionForGood, we use the persuasion strategy annotations as dialogue acts.} 
We annotate 50 utterances from each dialogue act in these datasets, using the automated GPT-4o annotation procedure described in \S~\ref{subsec:automatic-detection}. 

Figure~\ref{fig:friction-dialogacts-distribution} shows the distribution of utterances in each dialogue act that were annotated as each of the five friction super-categories (as well as ``No Friction''). 
We highlight several takeaways. 1) Almost all dialogue acts, across all three datasets, are expressed both with and without friction.
2) Dialogue acts comprising requests (``Request'' in MultiWOZ, ``Request Instruction'' and ``Request Object Loc.''\ in TEACh, ``Personal Inquiry'' in PersuasionForGood) are inherently frictive in nature, since they probe for information about the environment or the user's preferences. 
3) The most common forms of friction applied are ``Probing'' and ``Overspecification''. 
4) The prevalence of other friction categories depends on the dialogue data and task. 
Due to the embodied nature of TEACh, reflective pauses are more commonly observed. 
Similarly, several persuasion strategies in PersuasionForGood rely on revealing user assumptions and reinforcement.

%% file: sections_new/table_frictionexamples.tex
\begin{table*}[!t]
\centering
\resizebox{\linewidth}{!}{%
\begin{tabular}{@{}ll@{}}
\toprule
\textbf{Friction Movements}      & \textbf{Example Utterances}                                                                 \\ \midrule
Contextual Assumption Reveal     & “that’s the mug i think we have to use"                                                     \\
Conversational Assumption Reveal & "I assume you mean the center of town. We have many hotels in Cambridge."                   \\
Metacognitive Assumption Reveal  & “Yes, I think there's been some confusion.”                                                 \\ \midrule
Conversational Pause             & “hmm,” “...”, “Let me think,” “Let’s see,” “I’ll check now…”                                \\
Embodied Pause                   & {[}slowly approaches the target instead of directly grabbing{]}                             \\
Recalibrating Pause              & "Let's go back to lodgings for a moment.”                                                   \\ \midrule
Reinforcement &
  \begin{tabular}[c]{@{}l@{}}(Turn $t$) “Do you want a room for Thursday for 3 people, 2 nights?”  \\ (Turn $t+1$) “There are no guesthouses for 3 people for 2 nights starting on Thursday.” \\ (Turn $t+2$) “Should I book it for 3 people for 2 nights starting from Thursday?”\end{tabular} \\ \midrule
Elaborative Overspecification    & “i cleaned the mug.” (both interlocutors can see this) \\
Confirmative Overspecification   & “Good news! I was able to book two rooms for 5 nights at Finches B\&B for you.”        \\ \midrule
Contextual Probing               & "Which drawer should I open?”, "What area in Cambridge would you like to stay?" \\
Conversational Probing:          & “What did you say again?”, "You said you wanted tomatoes in your sandwich, right?"                                                                   \\
Plan-Level Probing               & “What's the next step I need to do?”, "Will we need this mug again later?"                                                                     \\ \bottomrule
\end{tabular}%
}
\caption{Examples from the MultiWOZ and TEACh datasets for all subcategories of friction movements. 
Subcategories can be extended or modified according to the specific conversational setting or dataset under consideration.}
\label{tab:friction_examples}
\vspace{-0.1in}
\end{table*}

%% file: sections_new/04_analyzing_friction.tex
\section{The Utility of Friction}
This section highlights how friction improves the modeling of users' mental states, while also highlighting its relationship to timing and utterance valence (\S\ref{sec:analyze_friction}). We also show how friction helps to accomplish user goals (\S\ref{sec:applying_friction}).

\subsection{A ``Valencing'' Act: Friction Helps Model User Mental States}
\label{sec:analyze_friction}
Previous analyses~(\S\ref{sec:comparative_analysis}) 
show friction movements can refine and extend existing taxonomies of dialogue utterances (e.g., acts or persuasion strategies). 
In this section, we provide an initial study to determine whether introducing friction into utterances can impact a model’s ability to infer mental states in task-oriented dialogue.

\paragraph{Experimental Setup} We focus on inferring user satisfaction in the MultiWOZ task \citep{eric2020multiwoz}, using annotations collected by \citet{sun2021simulating}. Tacitly, this task requires modeling the user's mental state regarding goal achievement.
 We use the average anticipated user satisfaction of the annotation cohort, which is a score on a 5-point scale. 
 For 1000 randomly sampled dialogues, we compute predictions of the user satisfaction for the conversation, using the method of \citet{sicilia2024evaluating}. This method prompts language models to infer the intensity of user beliefs on a continuous scale, producing state-of-the-art results on our current setting (i.e., user satisfaction in MultiWOZ).
 We report averaged results across GPT-4o, LLaMA-3.1 8B and 70B \citep{touvron2023llama}, and Mixtral 7x8Bv0.1 and 8x22B \citep{jiang2024mixtral}. 
 For each conversation, we also sample a random turn and annotate the friction movement using the automated procedure described in \S\ref{subsec:automatic-detection}.
 For each friction category, we also report the average turn number at which the sample happens and the average total dialogue length.
 
\paragraph{Hypothesis Testing} Our sampling strategy ensures independence of each data point, so unobserved turns are modeled as having a common (unobserved) effect on the prediction errors or other statistics. Specifically, we test the null hypothesis: \textit{the category of friction---including no friction---that occurs at any random turn does not impact the user modeling errors, length, etc.\ of the dialogue}. Under this null, regardless of unobserved turns, there should be no observed effects of a specific friction category across the whole dialogue. 
If that is not the case, we reject the null: some specific friction category, occurring at a random turn, has impact on the remaining dialogue.

 \begin{figure}[t]
    \centering
    \includegraphics[width=\columnwidth]{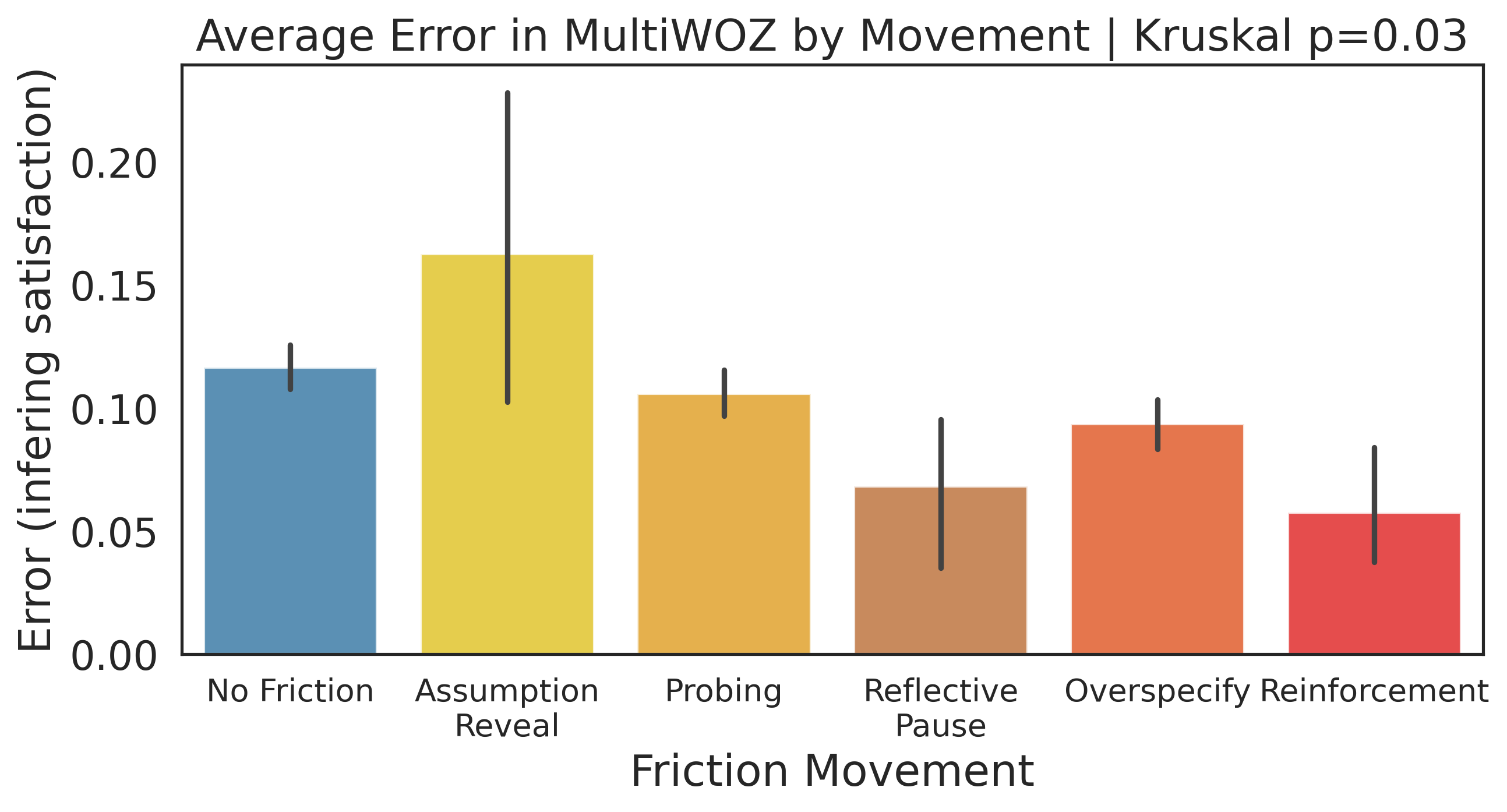}
    \caption{Mean squared error inferring user satisfaction from dialogue history within MultiWOZ. This task requires modeling user mental states. Kruskal-Wallis test for difference is significant. Visual inspection shows introducing friction reduces user modeling errors.}
    \label{fig:multiwoz-box}
\end{figure}

\paragraph{Friction Improves User Modeling} In Figure~\ref{fig:multiwoz-box}, we show bar plots of average squared model errors with 95\% confidence intervals when inferring user satisfaction grouped by friction movements. An interesting finding is that model errors---at inferring user satisfaction---tend to decrease for conversations when certain types of friction are identified. Indeed, a Kruskall-Wallis test for difference in error distribution (under the observation of friction) rejects the null that the distributions are the same. This result may be due to the nature of friction, which slows down the conversation to reveal more information about user goals or beliefs. 

\begin{figure}
    \centering
    \includegraphics[width=\columnwidth]{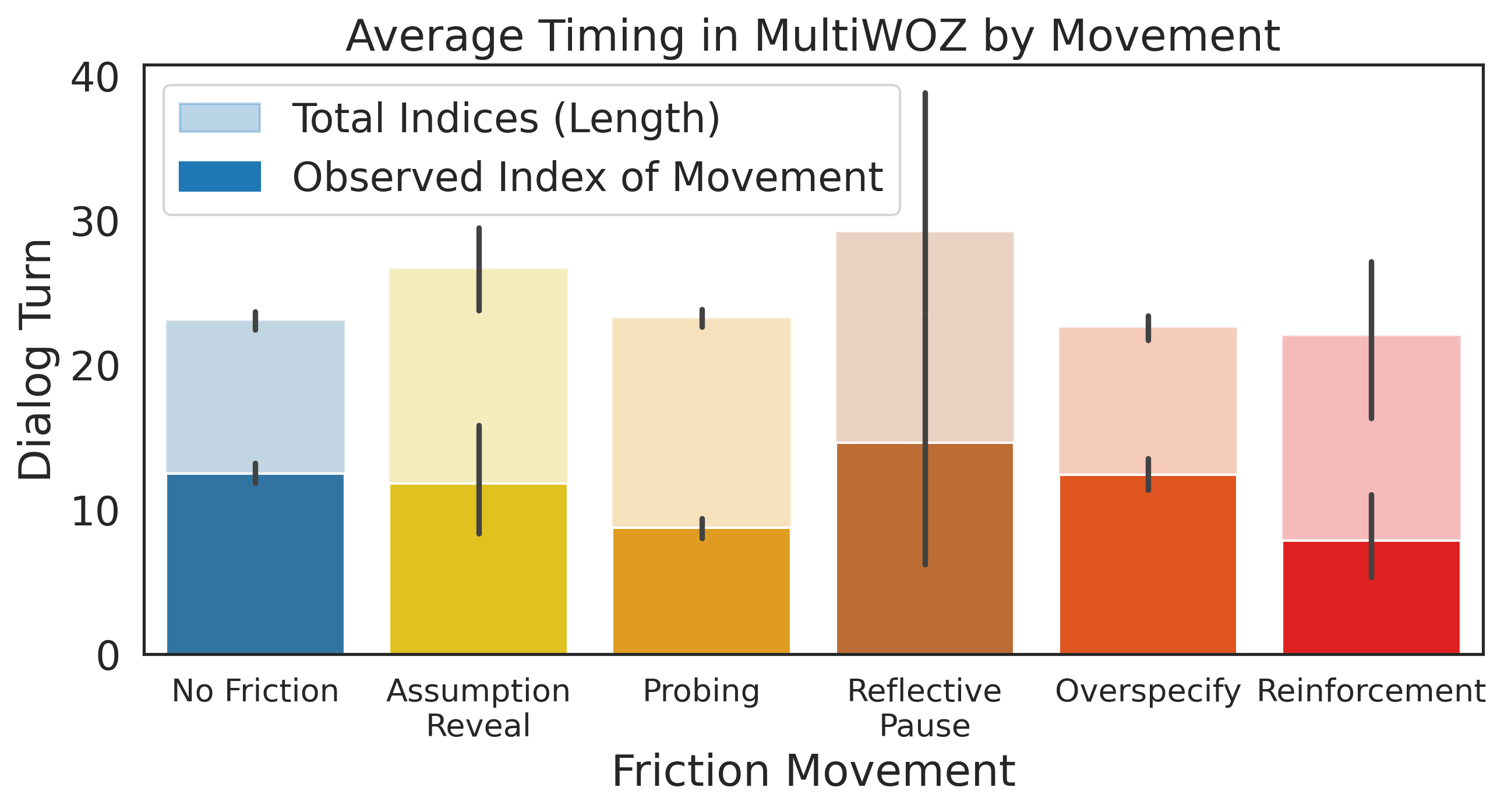}
    \caption{When each friction movement occurs (i.e., average index of observation) as well as average total dialogue length for each movement. Corpora consists of human ``Wizard of Oz'' data (i.e., MultiWOZ; \citealp{eric2020multiwoz}). Results show humans strategically use friction at different time points ($p<0.01$) and friction often ``slows down'' conversations ($p=0.1$).}
    \label{fig:multiwoz-timing-box}
\vspace{-0.1in}
\end{figure}

\paragraph{Friction Impacts Timing} Next, we address the hypothesis that friction movements slow down the dialogue, inducing higher valence and longer, more thoughtful conversations. In Figure~\ref{fig:multiwoz-timing-box}, we also show when (on average) different friction movements were observed, as well as the total average length of dialogues where we observed specific friction movements. Results indicate the relationship between friction and timing in human-human ``Wizard of Oz'' data. Use of friction movements (e.g., reveal and pause) tends to lengthen a dialogue or ``slow it down.'' Other friction categories (e.g., probing) tend to occur early in a dialogue, showing friction is used at strategic times.%

\paragraph{Takeaway} These results confirm our initial hypotheses about the utility of friction in inferring user mental states, its impact on dialogue length, and its strategic use in human conversation.

%% file: sections_new/05_applying_friction.tex
\subsection{A Guiding Force: Friction Helps To Accomplish User Goals}
\label{sec:applying_friction}
In this section, we apply the proposed friction taxonomy to goal-oriented conversational settings in a dynamic manner and study its impact on overall task success. Our experiments focus on two scenarios: (i) multi-domain booking agent, and (ii) embodied conversational system.

\input{sections_new/table_applyingfrictionresults}

\paragraph{Experimental Setup}
\label{sec:friction_multiwoz_alfworld_setup}
We utilize the widely adopted MultiWOZ~\cite{multiwoz} and ALFWorld~\cite{shridhar2020alfworld} datasets for our experiments on multi-domain booking and embodied scenarios, respectively. While both settings are text-based, they differ in the level of task embodiment. 
To generate dialogues for MultiWOZ, we employ AutoTOD~\cite{autotod} as the assistant model. For ALFWorld, we enhance the original ReAct framework with additional dialogue capabilities~\cite{respact} as there is no inherent dialogue. Additionally, for both cases, we leverage GPT-4o-mini as the user simulator to generate the user utterances when prompted with the dialogue level user goals and previous conversation context. In these experiments, we only consider a subset of high-level friction categories (assumption reveal, overspecification, and probing) based on the most frequently-occurring movements in these datasets as also shown in Figure~\ref{fig:friction-dialogacts-distribution}. Friction is introduced by adding definitions of friction categories and in-context examples in the LLM prompt.

The generations for both datasets are evaluated on task completion. For MultiWOZ, we adopt AutoTOD's~\cite{autotod} online \textit{Success} metric, where \textit{Success} is defined as the system's ability to identify the correct entity and provide all the attributes requested by the user. 
For ALFWorld, we use \textit{Task Success Rate}~\cite{shridhar2020alfred} where a task is successfully completed if, at the end of the action sequence, the objects are in the correct positions and states. 
We report the average task success over three runs to account for the variability in LLM responses. For more details, please refer to Appendix \S~\ref{sec:appendix_multiwoz} and \S\ref{sec:appendix_alfworld}.

\paragraph{Friction Improves Success on Multi-Domain Tasks} 
Table~\ref{tbl:mwoz_result} summarizes the results of incorporating friction into task-oriented conversations. We observe an improvement of approximately 3–6\% in task \textit{Success} for MultiWOZ. The system introduces assumption reveal and probing to clarify hidden assumptions and ambiguities, which contributes to improved task success. Overspecification, on the other hand, is typically introduced during the confirmation of an entity booking. In this case, the system explicitly restates all constraints provided by the user. This detailed specification assists the user in continuing the conversation in the correct direction, resulting in a higher performance than the other two categories. The model achieves the highest task success (62.8) when it incorporates all three friction categories, highlighting their combined effectiveness in improving task-oriented conversations. In this setup, the model generates a higher number of friction turns, with 56\% (33\% Probing and 23\% Overspecification) of the total turns containing friction.

\begin{figure}[!t]
    \centering
\includegraphics[width=.95\columnwidth]{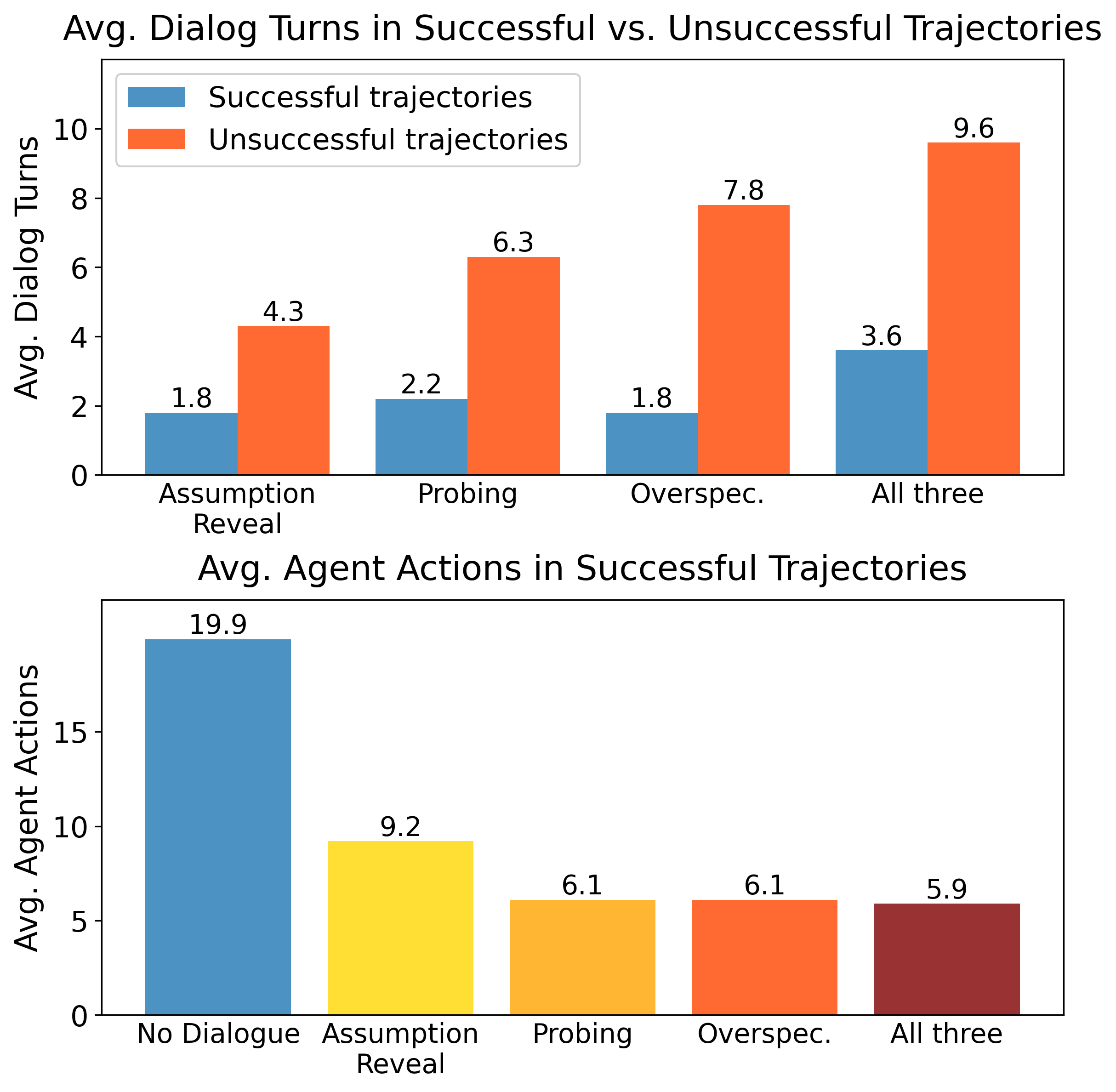}
    \caption{Impact of different friction strategies on task completion in ALFWorld. The figure quantifies the impact of dialogue as friction through turns and actions.}
    \label{fig:friction_embodied}
    \vspace{-0.1in}
\end{figure}

\begin{figure*}[!t]
    \centering
\includegraphics[width=\linewidth]{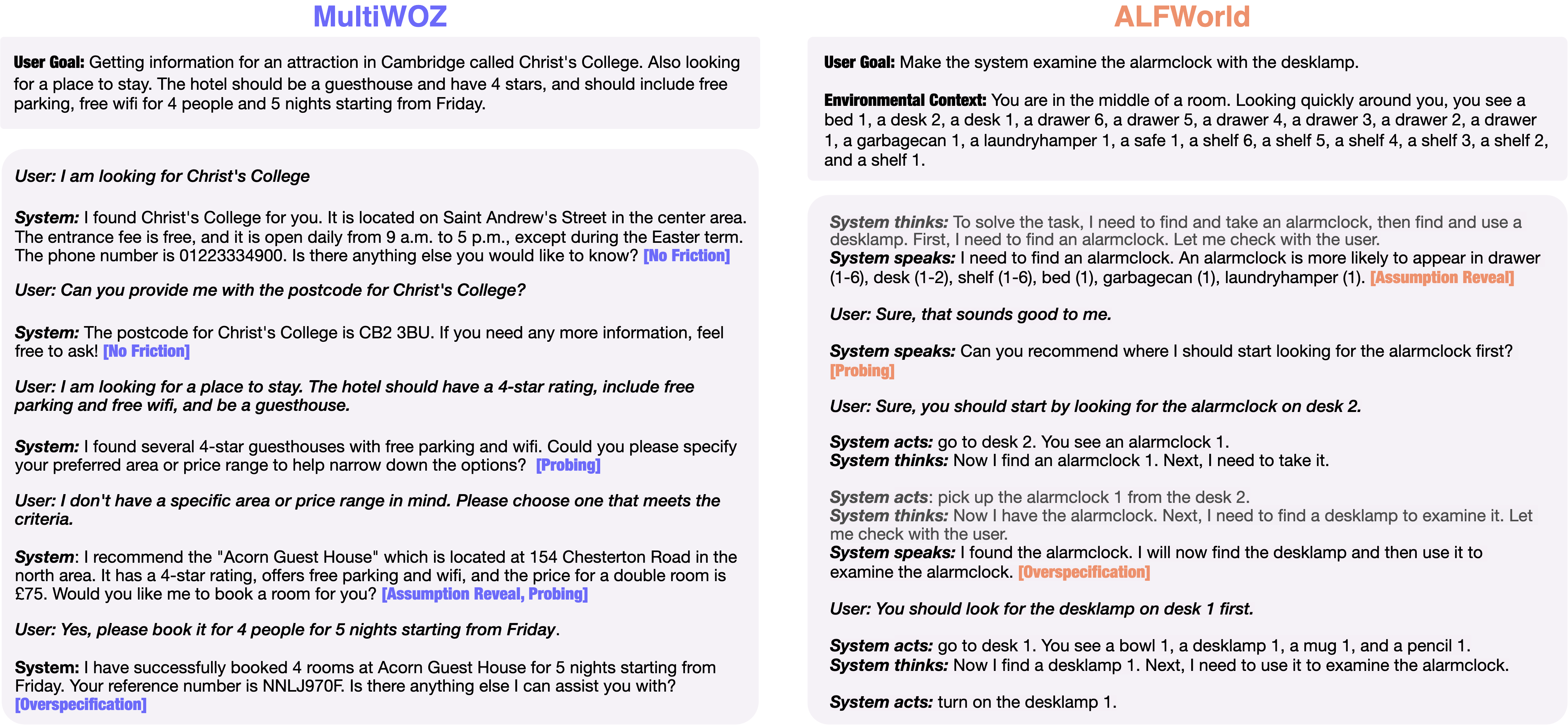}
    \vspace{-15pt}
    \caption{Examples of frictive dialogues in MultiWOZ and ALFWorld. These illustrate how friction impacts decision-making and execution in multi-domain booking agent \& text-based embodied conversational system}
    \label{fig:friction_examples}
    \vspace{-0.1in}
\end{figure*}

\paragraph{Friction Improves Success on Embodied Tasks}
In ALFWorld, we investigate different friction strategies in agent-user dialogue interactions. While any dialogue introduced by the agent inherently adds friction by temporarily pausing task execution, our results demonstrate that this friction can be beneficial. As shown in Table \ref{tbl:mwoz_result} and Figure \ref{fig:friction_embodied}, the baseline ``No Dialogue'' approach achieves a 51.49\% success rate but requires a high number of physical actions (19.9) as it progresses towards the goal without any user input. By introducing probing as a friction strategy, where the agent strategically pauses to ask clarification questions about task-critical information, we observe a significant improvement in task success (58.96\%). This approach not only increases the success rate but also substantially reduces the required physical actions to 6.1, while maintaining minimal dialogue overhead (2.2 turns in successful trajectories). These results demonstrate that introducing friction through strategic interactions can lead to more efficient task execution, despite the additional interaction time, by enabling better agent-user communication and task understanding. 

In ALFWorld with all three friction categories, we see a decrease in success rate compared to MultiWOZ.
We conjecture that this decrease is due to the step limit in the execution environment. When all three friction strategies are applied, they start consuming more steps for thinking and speaking (from 8.5 steps with Probing to 13.2 in All Three) instead of physical acts to enable the agent to progress toward task completion, resulting in a decrease in task success. 

\paragraph{Takeaway} For both MultiWOZ and ALFWorld, applying friction categories improves task success in goal-oriented conversations. The result demonstrates that incorporating friction can enhance the agent's understanding of user goals, resulting in higher task success. 
Qualitative examples of frictive conversations are provided in Figure~\ref{fig:friction_embodied}.

%% file: sections_new/table_applyingfrictionresults.tex
\begin{table*}
\centering
\begin{small}
\begin{tabular}{lrrrrrrr}
    \toprule
    \multirow{4}{*}{\begin{tabular}{l} \textbf{Friction} \\ \textbf{Movement}\end{tabular}} & \multicolumn{3}{c}{\textbf{MultiWOZ}} & & \multicolumn{3}{c}{\textbf{ALFWorld}} \\ 
    \cmidrule{2-4} \cmidrule{6-8}
\centering 
& \textbf{Success} & \begin{tabular}{c} \textbf{Fric.} \\ \textbf{(\%)}\end{tabular} & \begin{tabular}{c} \textbf{Avg.} \\ \textbf{Turns}\end{tabular}& & \textbf{Success} & \begin{tabular}{c} \textbf{Fric.} \\ \textbf{(\%)}\end{tabular} & \begin{tabular}{c} \textbf{Avg.} \\ \textbf{Turns}\end{tabular} \\ \midrule
No Friction & 56.40 \scriptsize{$\pm$ 2.3}  & 0.0 & 4.8 & & 51.49 \scriptsize{$\pm$ 0.8} & 0.0 & 0.0 \\ 
Assumption Reveal & 59.00 \scriptsize{$\pm$ 1.3}  & 24.9 & 4.5 & & 52.18 \scriptsize{$\pm$ 0.9} & 12.2 & 6.1\\ 
Probing & 59.87 \scriptsize{$\pm$ 0.8}  & 31.2 & 4.6 & & \textbf{58.96 \scriptsize{$\pm$ 2.2}} & 17.0 & 8.5\\ 
Overspecification & 61.93 \scriptsize{$\pm$ 0.8}  & 24.3 & 4.7 & & 52.93 \scriptsize{$\pm$ 2.8}& 19.6 & 9.8\\ 
All three & \textbf{62.80} \scriptsize{$\pm$ 1.3}  & 56.1 & 4.8 & & 46.06 \scriptsize{$\pm$ 1.3}& 26.4& 13.2\\ 
\bottomrule
\end{tabular}
\end{small}
\caption{\label{tbl:mwoz_result} 
The table illustrates the impact of introducing friction on the overall task success of goal-oriented conversations from MultiWOZ and ALFWorld test data. \textit{Success} indicates the fraction of conversations where the system satisfies all the user requirements. In MultiWOZ, ``None'' refers to the AutoTOD baseline (without friction), while in ALFWorld, it denotes the ReAct baseline (actions only, no dialogue).
}
\vspace{-0.1in}
\end{table*}

%% file: sections_new/06_discussion.tex
\subsection{Discussion: Open Questions \& Implications}
\label{sec:discussion}
We now discuss some important questions for synthesizing a path forward for the use of friction in conversational system design.

\paragraph{Are all questions friction?} As seen in the results of Table~\ref{tbl:mwoz_result} and Figure~\ref{fig:friction_embodied}, probing brings major gains in task success. In addition, Figure~\ref{fig:friction-dialogacts-distribution} shows that question-based dialogue acts such as \texttt{Request} or \texttt{Confirm} are almost always mapped to \textit{Probing}. This raises the question of whether all questions are inherently frictive since the interlocutor could always proceed without asking by making assumptions instead. 
A qualitative analysis of our annotations reveals that questions are also used in non-frictive ways (e.g. ``Want to try another option?'') that are intended to move the conversation forward. 
Further, we also see questions used to achieve means other than probing, such as revealing assumptions (``I think the mug is clean?''). 
Hence, current data shows that not all questions are friction, but qualitatively, humans prefer to ask questions in frictive moments. This nuanced perspective on how humans ask questions suggests more work is needed to study how dialogue systems represent uncertainty and ask questions. Our taxonomy provides a useful characterization of probing behavior for this purpose.

\paragraph{Are all utterances friction?} The results in Figure~\ref{fig:friction_examples} show that introducing any utterance or friction movement significantly reduces physical actions. Cognitive science literature supports this finding, suggesting that conversations reduce physical acts while fostering collaboration. Then, can any utterance be considered positive friction? While they may be in embodied settings—where utterances inherently add turns—our MultiWOZ experiments indicate that not all utterances qualify as friction in text-only domains. In embodied interactions, utterances disrupt action flow, emphasizing the dual nature of friction as both a turn-level and discourse-level phenomenon. Whether an utterance is considered friction ultimately depends on its context as an interruption, underscoring the need for dialogue policies that address both turn- and interaction-level dynamics.

\paragraph{How can friction be incorporated into reward mechanisms?} 
Preference ratings for RLHF datasets are typically not collected over multi-step human-LM interactions. 
Consequently, reward models can optimize short-term user satisfaction over longer-horizon collaboration goals, and thus friction movements are not naturally built into RL-trained dialogue policies. 
We believe our taxonomy can be used as meta-labels, providing scaffolding for both collecting preference ratings with an emphasis on utterance valence (as demonstrated in~\S\ref{sec:analyze_friction}), and for designing reward models that emphasize collaboratively building common ground over multiple interactions.

\paragraph{When does friction become negative?}
In this work, we advocate for momentarily slowing down dialogues to achieve better long-horizon outcomes. 
However, introducing too much friction (in the form of too much reflection or probing without taking any actions) can increase user frustration and disengagement, having a negative impact in the long term. 
Therefore, evaluating frictive movements in LLMs necessitates new evaluation paradigms that adequately balance short-term efficiency and utterance valence with long-term task completion. 
Further, each user has different requirements in the efficiency vs. efficacy trade-off, and thus the amount of friction introduced could be personalized to the observed preferences of the user.

\section{Conclusion}
Our taxonomy and experiments reveal that incrementally building common ground via positive friction is beneficial for goal-oriented dialogue, both in terms of improving task success (\S\ref{sec:applying_friction}) and modeling user mental states (\S\ref{sec:analyze_friction}).
Overall, we observe that considering friction is a fresh and fluid perspective for dialogue systems that builds on theories of cognition, discourse, and dialogue. Friction has future implications in addressing the valence disparity in dialogue management policies, and allowing users to reflect on their own reasoning instead of relying on the generations of frictionless LLMs. We further plan to investigate the optimal timing and application of different types of friction in goal-oriented dialogues. 

%% file: sections_new/07_trailing_sections.tex
\clearpage

\section*{Ethics Statement}
Our study contains two main points of ethical consideration: use of human subjects during annotation of friction, and use of LLMs in automatic friction detection. We have followed the guidelines of our institution's IRB protocols during the recruitment and administration of the annotations. Further, we use closed-source LLMs (e.g. GPT-4o) in our experimental setups and automatic friction detection, and we acknowledge that these models may perpetuate biases in their training data that is unknown to the public. As our datasets do not contain controversial utterances or emotionally burdening topics, we do not anticipate bias creation and laundering due to our use of LLMs.

\section*{Limitations}
We have evaluated the utility of our friction strategies using traditional metrics such as user satisfaction and task success. However, this new lens of including more positive instances of friction into dialogue systems to better model user mental states opens up avenues for developing new generation metrics focused on it. An important aspect of evaluation is the distinction between short-term versus long-term goals (as defined in \S\ref{sec:pos_fric}) and values and how to measure them in a conversation. In addition to traditional metrics of number of turns, lexical diversity, final task success, we posit that it is important to maximize long-term goals over short-term gains. These can be defined specific to each dataset, and need to be evaluated accordingly. For instance, in the embodied setting in \S\ref{sec:applying_friction}, we used the count of API calls to physical actions and the reduction of them as a way of measuring the effects of friction. 

\section{Acknowledgements}
This research was supported in part by Other
Transaction award HR0011249XXX from the
U.S. Defense Advanced Research Projects Agency
(DARPA) Friction for Accountability in Conversational Transactions (FACT) program. Researchers from UIUC have benefited from the Microsoft Accelerate Foundation Models Research (AFMR) grant program, through which leading foundation models hosted by Microsoft Azure and access to Azure credits were provided to conduct some parts of this research. Further, Tejas Srinivasan completed parts of this work while being supported by the Amazon ML Fellowship from the USC-Amazon Center for Secure and Trusted Machine Learning. We want to thank Weiyan Shi for her contributions and guidance. In addition, we thank Kate Atwell and Saki Imai for their helpful feedback.

%% file: sections_new/xx_appendix.tex
\section{Annotation Interface}
\label{sec:appendix_annotation_interface}
\begin{figure}[!h]
    \centering
    \begin{tabular}{cc}
        \includegraphics[width=0.45\linewidth]{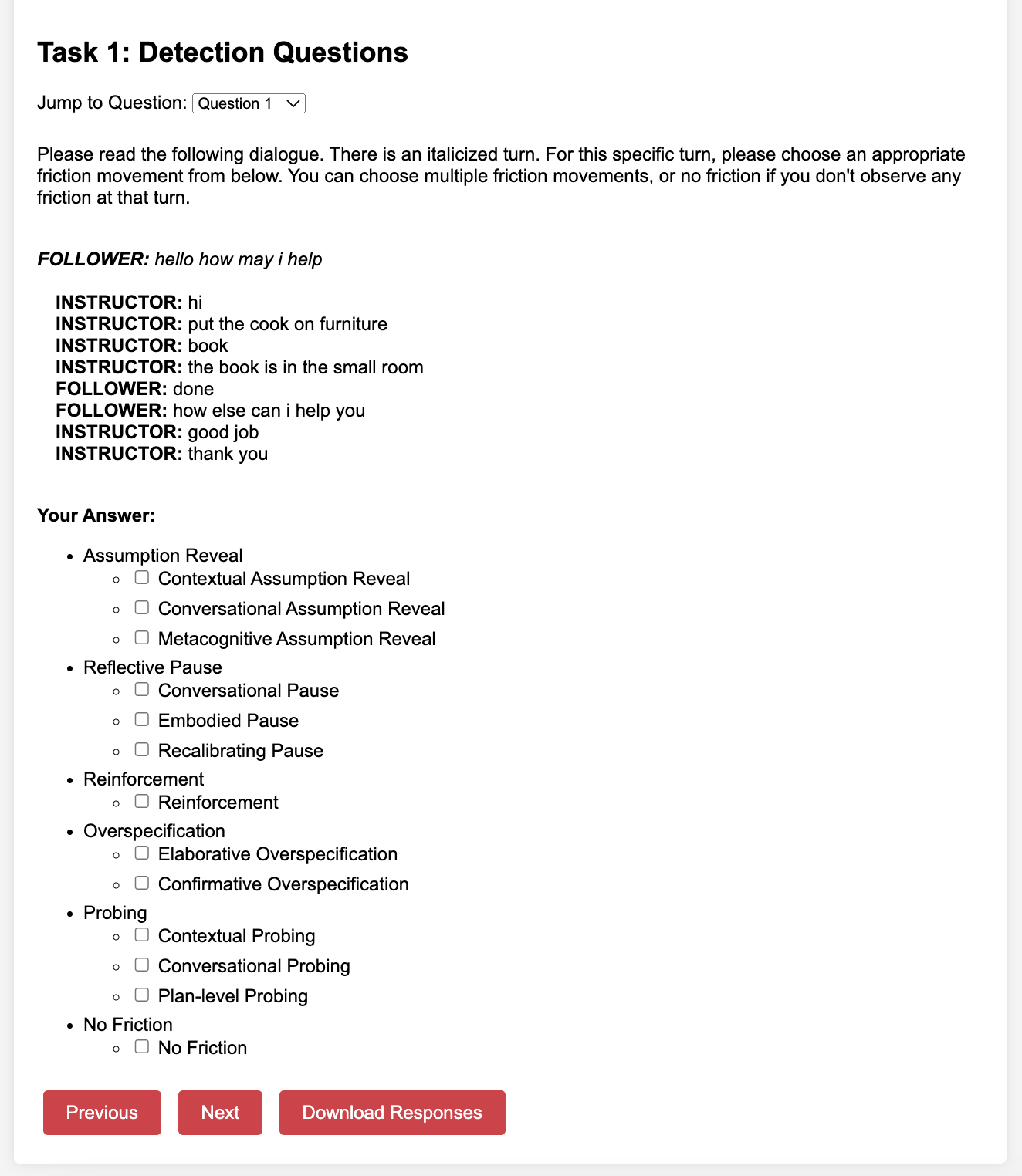} & \includegraphics[width=0.45\linewidth]{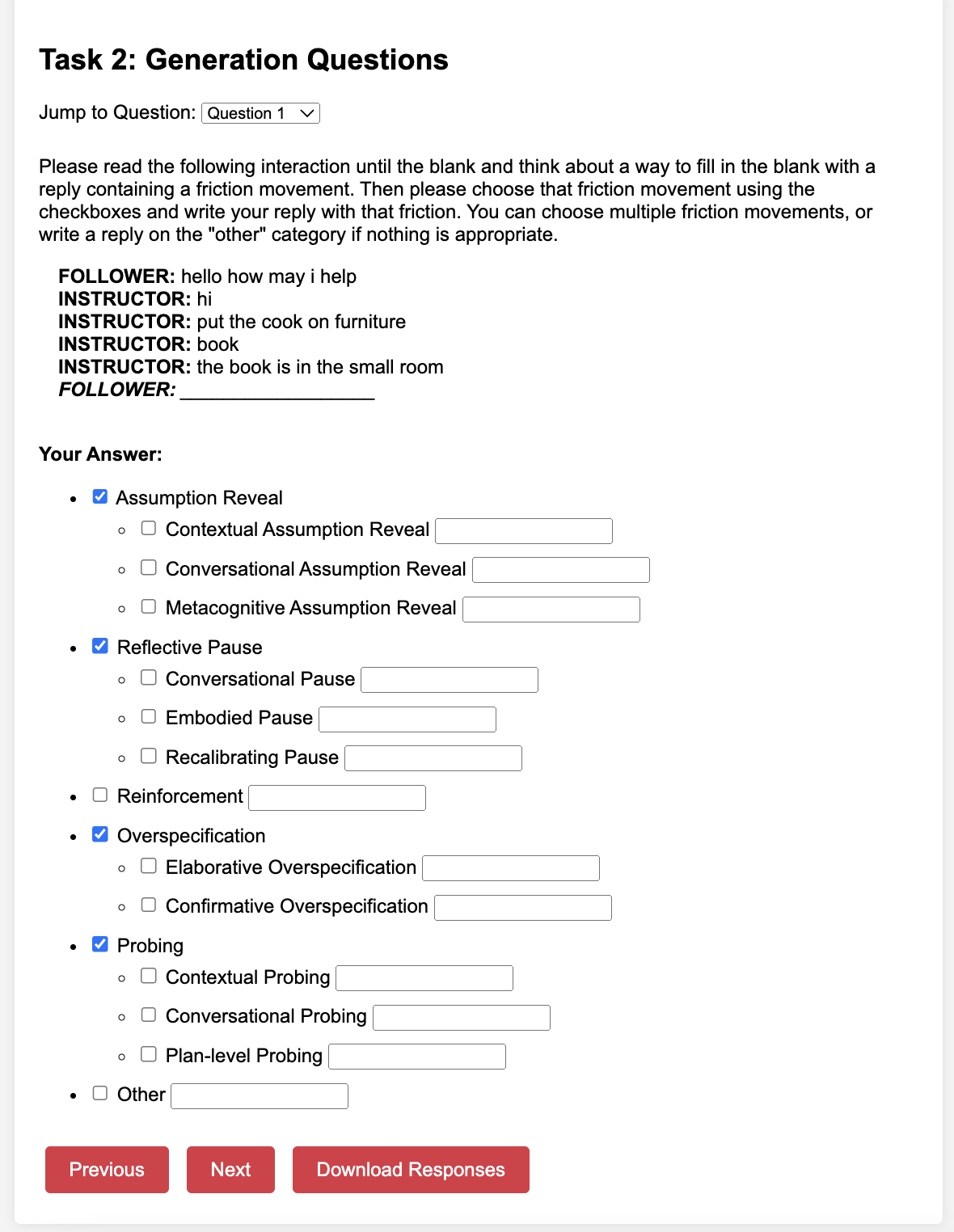}
    \end{tabular}
    \caption{This figure shows the annotation interface used for the human data collection of this study. Left corresponds to the detection task while the right one is the production task.}
    \label{fig:annotation_interface}
\end{figure}

\section{Inter-Annotator Agreement Heatmaps}
\label{sec:appendix_interannotator}

\begin{figure}[!h]
    \centering
    \begin{tabular}{cc}
         \includegraphics[width=0.5\linewidth]{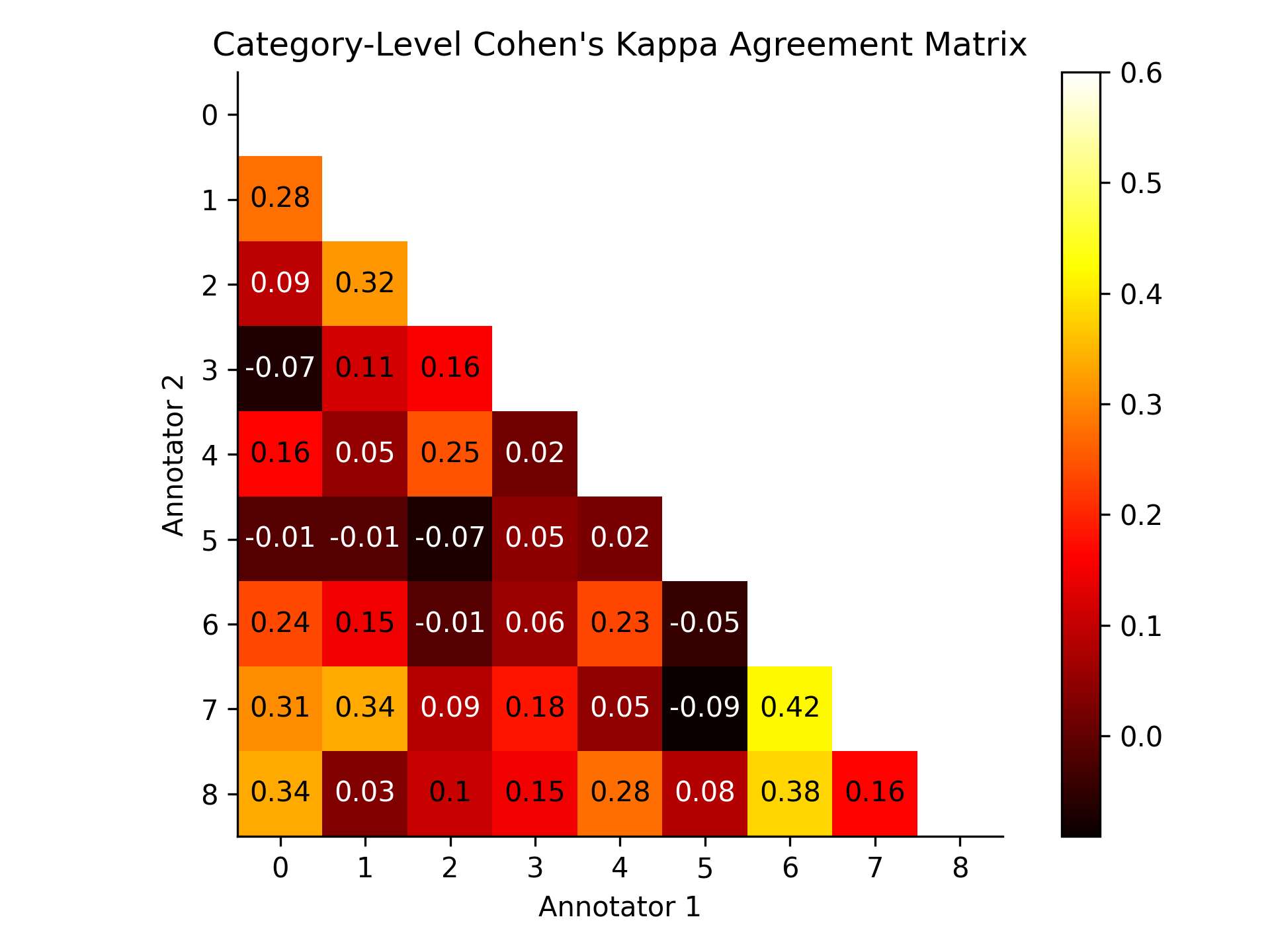} &  \includegraphics[width=0.5\linewidth]{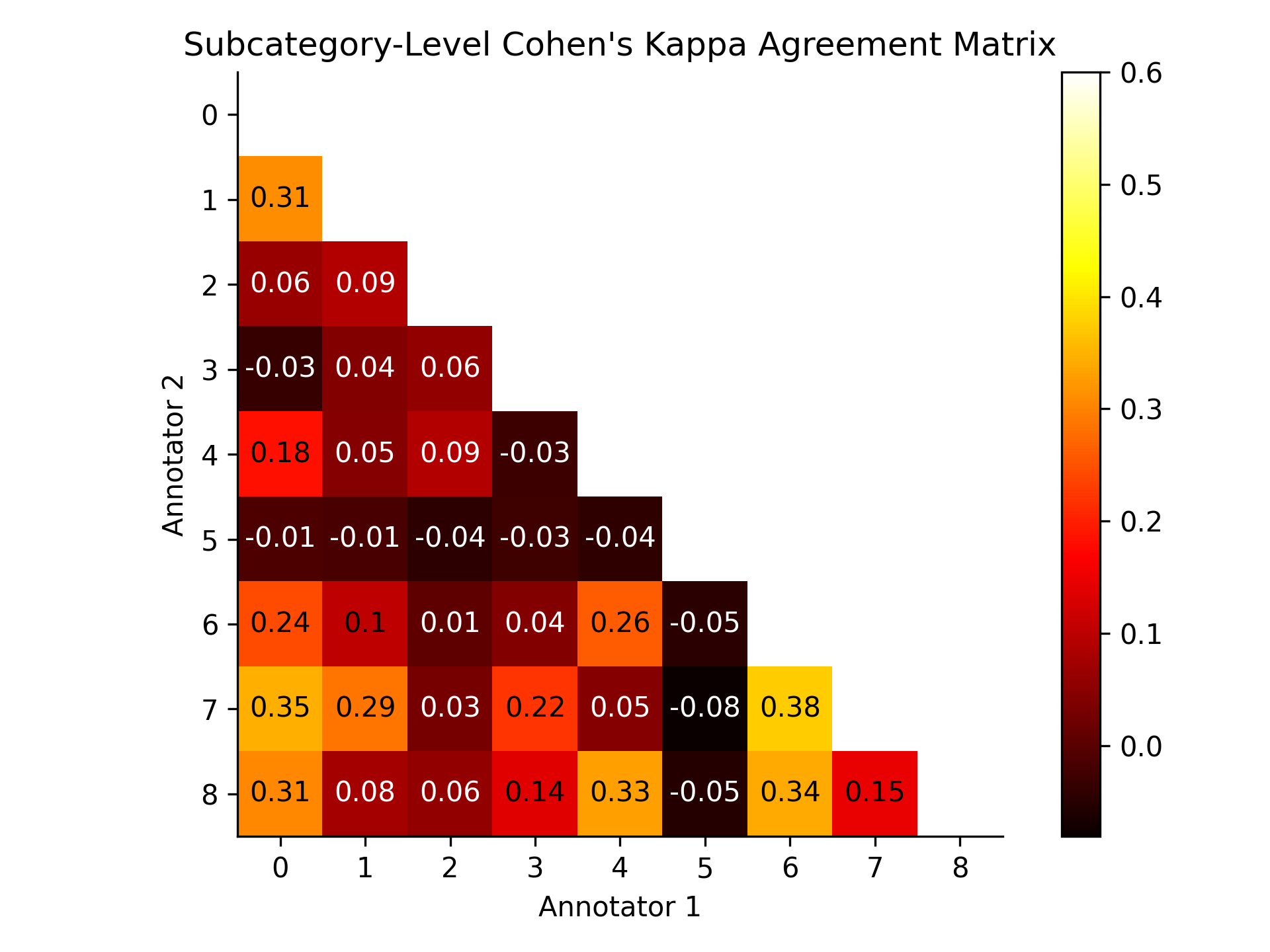}
    \end{tabular}
    
    \caption{This figure shows the category and subcategory level Cohen's kappa agreement scores for all the annotators. The highest measurements are between annotators 6 and 7. Both axes correspond to annotator IDs.}
    \label{fig:subcategory_interanno}
\end{figure}

\clearpage
\section{Histograms of Friction Movements in Human Annotation}
\label{sec:appendix_histogram}

\begin{figure}[!h]
    \centering
    \begin{tabular}{cc}
       \includegraphics[width=.45\linewidth]{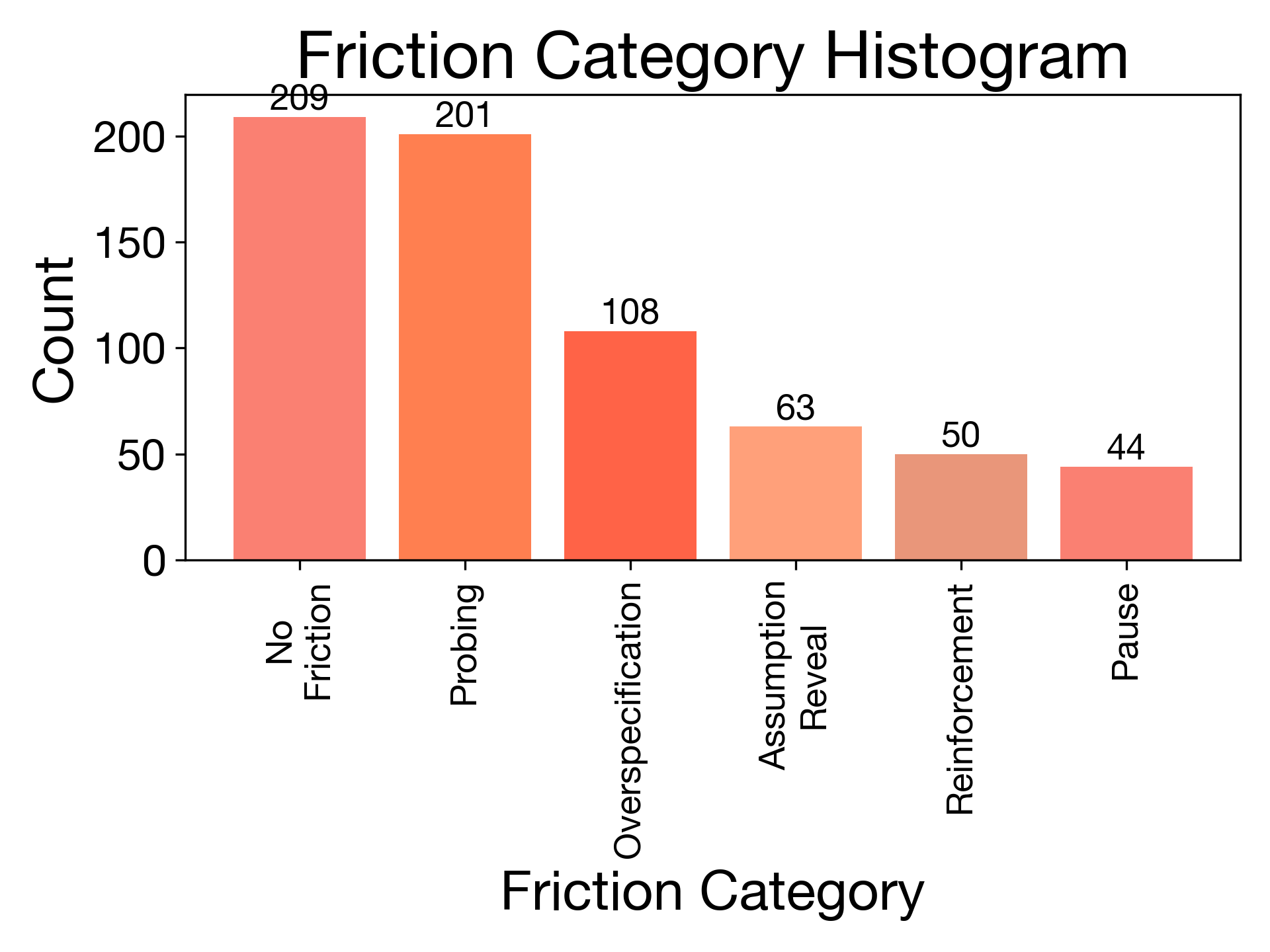}  &  \includegraphics[width=.45\linewidth]{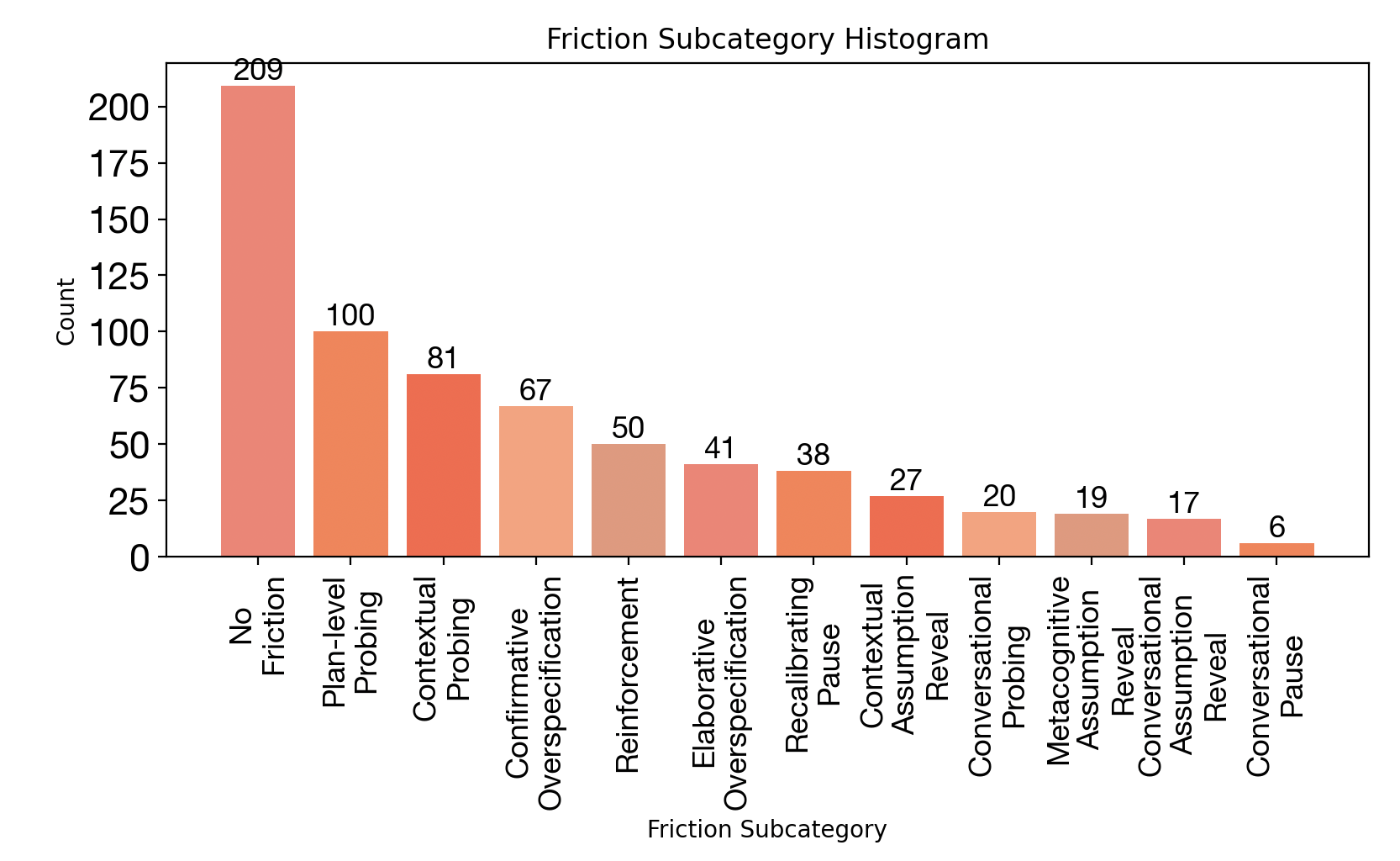}
    \end{tabular}
    \caption{Histograms of friction movement categories and subcategories for the collected friction detection annotations.}
    \label{fig:histograms}
\end{figure}

\begin{figure}[!h]
    \centering
    \begin{tabular}{cc}
       \includegraphics[width=.45\linewidth]{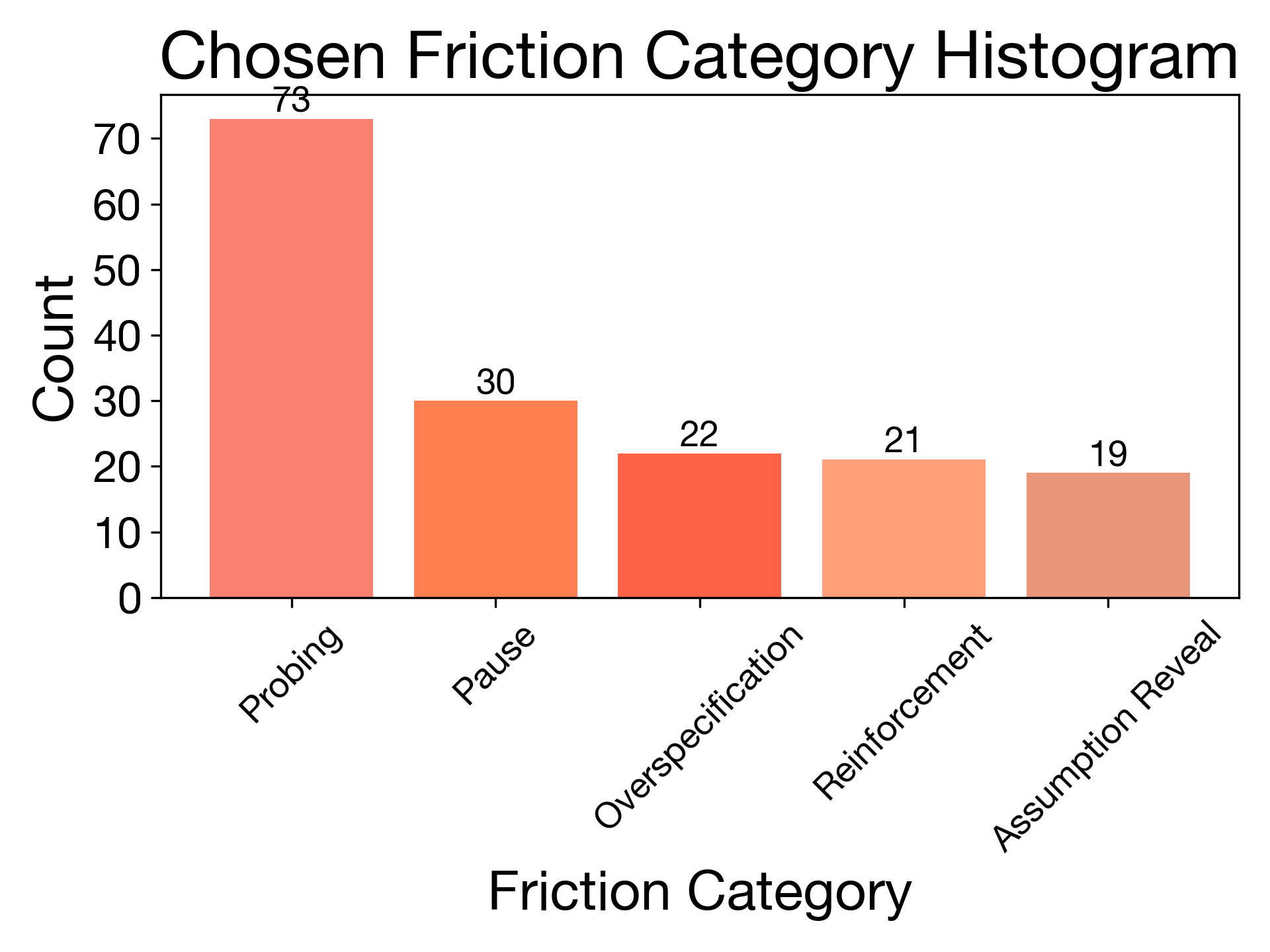}  &  \includegraphics[width=.45\linewidth]{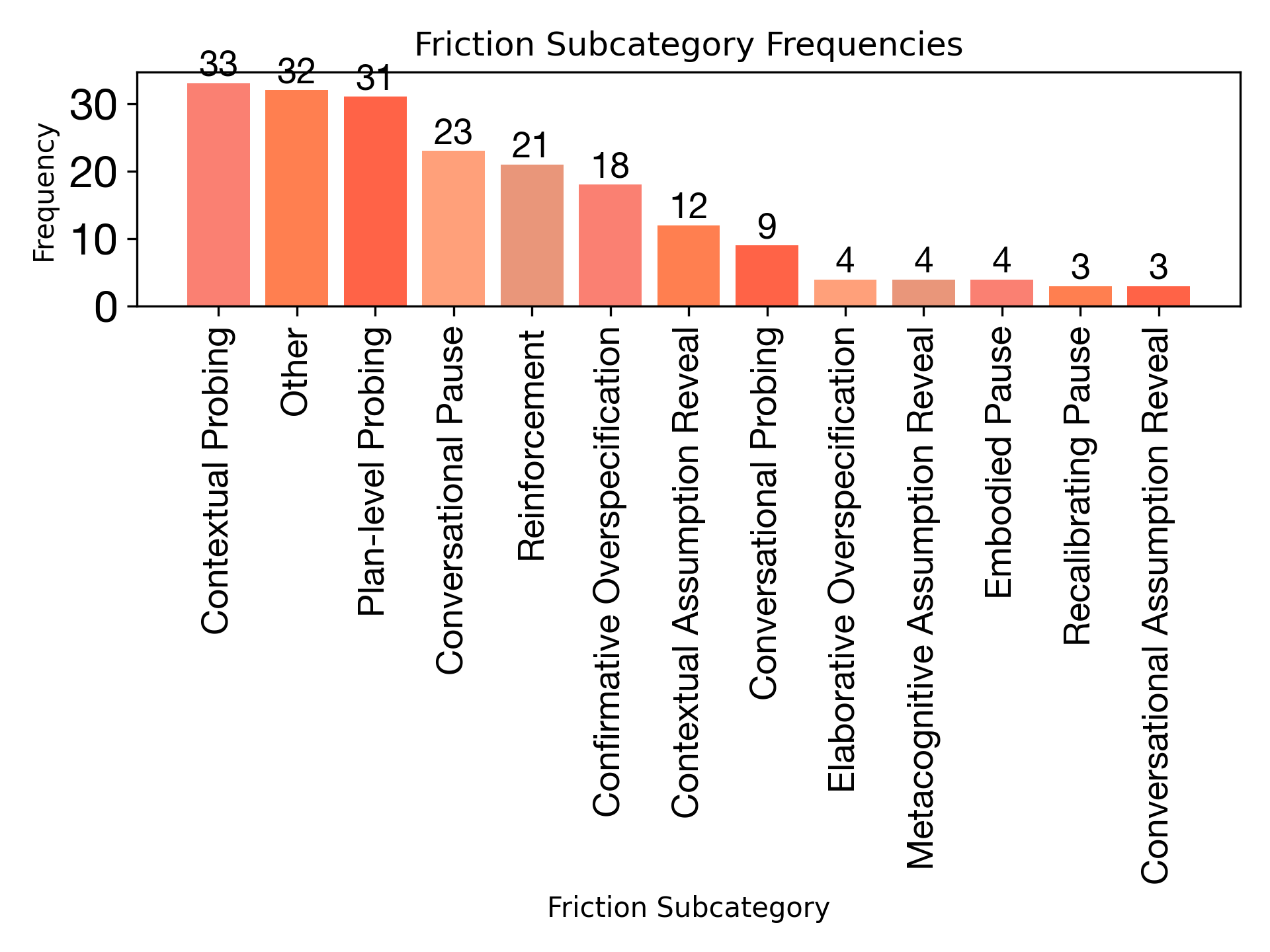}
    \end{tabular}
    \caption{Histograms of friction movement categories and subcategories for the collected friction production annotations.}
    \label{fig:histograms}
\end{figure}

\section{Detailed Analysis of Automatic Detection of Friction}
\label{sec:appendix_in_out_gpt}

\begin{figure}[!h]
    \centering
\includegraphics[width=.8\linewidth]{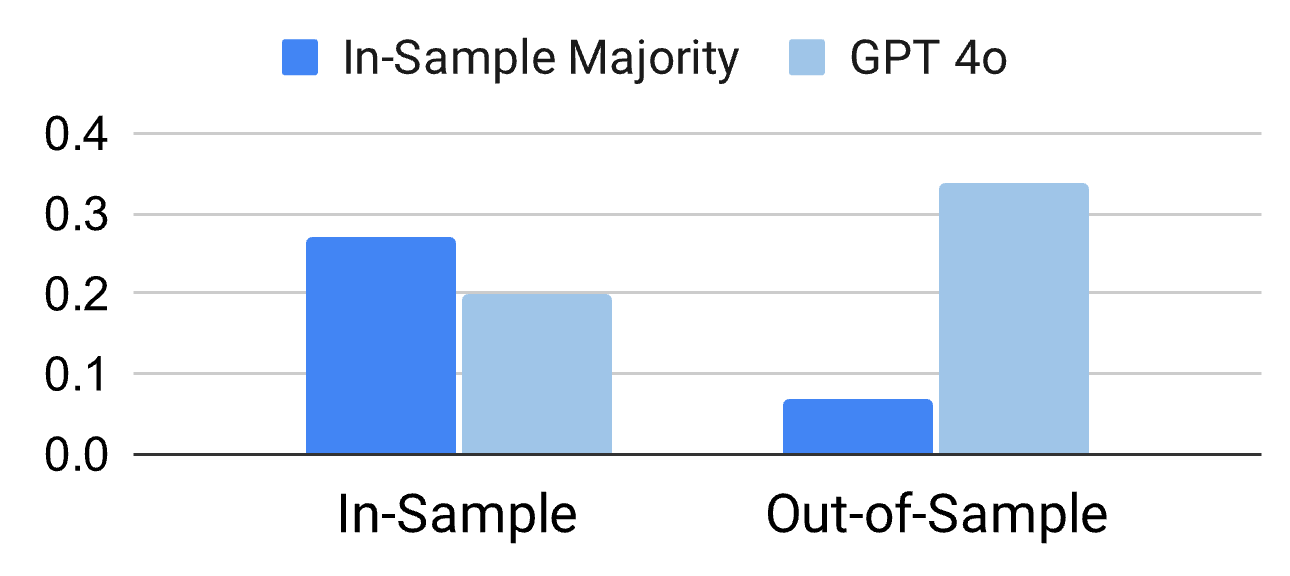}
    \caption{Cohen's Kappa for GPT-4o and in-sample majority vote, averaged across different voter groups. On average, GPT-4o agrees with out-of-sample voters more, demonstrating that automated annotation may better represent diverse opinions outside of the collected annotations.}
    \label{fig:gpt_v_maj}
\end{figure}
Overall, these agreements are more consistent than those observed among human annotators. Figure~\ref{fig:gpt_v_maj} also compares GPT-4o with a majority vote, focusing specifically on how well both annotation methods represent the opinion of annotators \textit{outside} of the sample used to compute the majority. Specifically, we use only 5 of 9 annotators to compute the majority check performance on the held-out remainder. As expected, a majority vote is the best representation of voters within a sample. Meanwhile, the majority vote may not be representative of other populations outside the sample used to compute the majority. In these cases, the automated assignments given by GPT-4o appear to be more representative, suggesting the subjective nature of the task may cause traditional majority vote annotation to be over-fit. We use GPT-4o to assign friction categories in the remainder of this work, since it is most cost-effective and since the human majority vote has limited robustness with respect to the choice in annotators.

\section{Details of MultiWOZ \& AutoTOD Setup}
\label{sec:appendix_multiwoz}
The MultiWOZ dataset comprises task-oriented conversations in which users interact with an agent to book hotels, restaurants, attractions, trains, and taxis. The task involves modeling a conversational agent that understands the user's goal and takes necessary actions to complete the booking. Let $D_{t}=\{(U_0), (S_1, U_1), ... (S_t, U_t) \}$ be the dialogue history till turn $t$ where $S_i$ and $U_i$ be the system and user utterance at turn $i$, respectively. The task of the dialogue system is to generate $S_{t+1}$ after each turn $t$ such that it helps the user to complete the user goal. In this work, we incorporate friction into AutoTOD~\cite{autotod}, a state-of-the-art task-oriented dialogue generation model. AutoTOD utilizes the ReAct~\cite{react} framework, incorporating an instruction schema that integrates task descriptions and external APIs, enabling the system to automatically determine the appropriate action and generate the system responses. Let $P$ represent AutoTOD's prompt for generating the system response $S_{t+1}$. To incorporate the friction classes, we modify $P$ to $P_\textsubscript{friction}$, which includes the definition of the friction class alongside an in-context example for each class. We use GPT-4o-mini as the LLM backbone and use temperature 0 for all the generations.

For MultiWOZ, task completion is traditionally evaluated using two metrics - Inform and Success~\cite{multiwoz}. The Inform metric evaluates whether the system identifies the correct entity for the user. The Success metric is stricter than Inform which determines whether the system provides all the required attributes for the identified entity. Since we are introducing friction turns, a direct comparison of the generated and ground-truth response is not possible. This is why we adopt AutoTOD's~\cite{autotod} online version of Success. The evaluation leverages GPT-4o-mini to check if all the user goals have been accomplished given the dialogue context through a question-answering task. Since the process involves GPT-4 call, we take an average of 3 runs to report the final \textit{Success} metric. The online evaluation setup requires a user simulator to generate the next user utterance. We use GPT-4o-mini as our user simulator that takes the user's goals and dialogue history. and generates $U_{t+1}$. The goal and the initial user simulator utterance ($U_0$) are directly taken from the MultiWOZ test dataset. The conversation concludes when the user simulator determines that the goal has been achieved and produces a special termination signal. The experiments are performed on 100 randomly selected conversations from the MultiWOZ test data.

\section{Details of the Alfworld Setup}
\label{sec:appendix_alfworld}
ALFWorld is a simulated environment based on the TextWorld framework~\cite{cote2019textworld} and aligned with the embodied ALFRED benchmark~\cite{shridhar2020alfred}. It provides a text-based interface for interacting with various physical tasks. ALFWorld comprises six categories of tasks, including finding hidden objects (e.g., locating a key inside a cabinet), moving objects (e.g., placing a cup on a table), manipulating objects with other objects (e.g., heating a potato in a microwave), and examining objects (e.g., inspecting a book under a desklamp). Each task instance in ALFWorld consists of more than 50 locations and requires an expert policy more than 50 steps to solve the task. Thus, the task requires understanding the environment, executing multi-step plans, and maintaining task-relevant state information. In our experiments we evaluate on \textbf{134} unseen evaluation games from the dataset.

To model the dialogue agent, we extend the original ReAct\cite{react} setup with additional dialogue capabilities for the agent~\cite{respact}. It is important to note that the original setup does not include any dialogues and involves only actions. This is why the extension is necessary because we introduce friction through dialogues. The agent can ask contextually relevant questions (e.g., "Where should I search for the knife in the kitchen?") and seek information effectively. 

Assume that an embodied agent operates in this environment. At time step $t$, it receives an observation $o_t$ from the environment, where $o_t \in \mathcal{O}$, where $\mathcal{O}$ represents the observation space. The agent executes an action $a_t \in \mathcal{A}$. Ideally, the agent's decision-making is based on a policy $\pi: \mathcal{C} \rightarrow \mathcal{A}$ where $\mathcal{C}$ where C represents the context space. The context $c_t$ encapsulates the relevant information available to the agent at time step $t$, including the current observation and the history of previous observations and actions: $c_t = (o_1, a_1, \cdots, o_{t-1}, a_{t-1}, o_t)$.

In our environment, the embodied agent operates as follows: At each time step \( t \), the agent receives an observation \( o_t \) from the environment, where \( o_t \in \mathcal{O} \), with \( \mathcal{O} \) representing the space of possible observations. Based on this, the agent executes an action \( a_t \in \mathcal{A} \), where \( \mathcal{A} \) is the action space.

The agent's behavior is governed by a policy \( \pi: \mathcal{C} \rightarrow \mathcal{A} \), where \( \mathcal{C} \) denotes the context space. The context \( c_t \) encapsulates all relevant information available at time step \( t \), including the current observation and the history of prior observations and actions. Formally, this is represented as:  
\[ c_t = (o_1, a_1, o_2, a_2, \dots, o_{t-1}, a_{t-1}, o_t). \]